\documentclass[sigconf]{acmart}

\usepackage{booktabs} 
\usepackage{multirow}
\usepackage{diagbox}
\usepackage{makecell}

\usepackage{flushend}
\usepackage{hyperref}
\usepackage{dsfont,amsfonts,amssymb,amsmath,color}
\usepackage{latexsym}
\usepackage{amsmath}
\usepackage{amsfonts}
\usepackage{graphicx}
\usepackage{subfigure}
\usepackage{algorithm} 
\usepackage[noend]{algpseudocode}
\usepackage{stfloats}
\usepackage{float}

\def\BibTeX{{\rm B\kern-.05em{\sc i\kern-.025em b}\kern-.08emT\kern-.1667em\lower.7ex\hbox{E}\kern-.125emX}}

\usepackage{balance}

\copyrightyear{2019} 
\acmYear{2019} 
\acmConference[CIKM '19]{The 28th ACM International Conference on Information and Knowledge Management}{November 3--7, 2019}{Beijing, China}
\acmBooktitle{The 28th ACM International Conference on Information and Knowledge Management (CIKM '19), November 3--7, 2019, Beijing, China}
\acmPrice{15.00}
\acmDOI{10.1145/3357384.3358061}
\acmISBN{978-1-4503-6976-3/19/11}

\begin{document}
\fancyhead{}

\title[HeteSpaceyWalk: A Heterogeneous Spacey Random Walk Framework for HIN Embedding]{HeteSpaceyWalk: A Heterogeneous Spacey Random Walk for Heterogeneous Information Network Embedding}

%







\author{Yu He$^{1,2}$ and Yangqiu Song$^3$ and Jianxin Li$^{1,2}$ and Cheng Ji$^{1,2}$ and Jian Peng$^4$ and Hao Peng$^{1,2}$}
\affiliation{%
  \institution{$^1$ Beijing Advanced Innovation Center for Big Data and Brain Computing, China\\$^2$ SKLSDE Lab, Beihang University, China\\$^3$ Department CSE, HKUST\\ $^4$Department of CS, UIUC}
}
\email{{heyu,lijx,jicheng,penghao}@act.buaa.edu.cn,  yqsong@cse.ust.hk, jianpeng@illinois.edu}

\renewcommand{\shortauthors}{Y. He et al.}

%
\begin{abstract}
Heterogeneous information network (HIN) embedding has gained increasing interests recently.
However, the current way of random-walk based HIN embedding methods have paid few attention to the higher-order Markov chain nature of meta-path guided random walks, especially to the
stationarity issue.
In this paper, we systematically formalize the meta-path guided random walk as a higher-order Markov chain process,
and present a heterogeneous personalized spacey random walk to efficiently and effectively attain the expected stationary distribution among nodes.
Then we propose a generalized scalable framework to leverage the heterogeneous personalized spacey random walk to learn embeddings for multiple types of nodes in an HIN guided by a meta-path, a meta-graph, and a meta-schema respectively.
We conduct extensive experiments in several heterogeneous networks and demonstrate that our methods substantially outperform the existing state-of-the-art network embedding algorithms.
\end{abstract}

%
%


%

%

%
\maketitle
\pagestyle{empty}  
\thispagestyle{empty} 
\section{Introduction}\label{sec:introduction}

\begin{figure}
	\small
	\setlength{\abovecaptionskip}{-0.1cm}
	\setlength{\belowcaptionskip}{-0.5cm}
	\centering
	\includegraphics[width=0.5\textwidth]{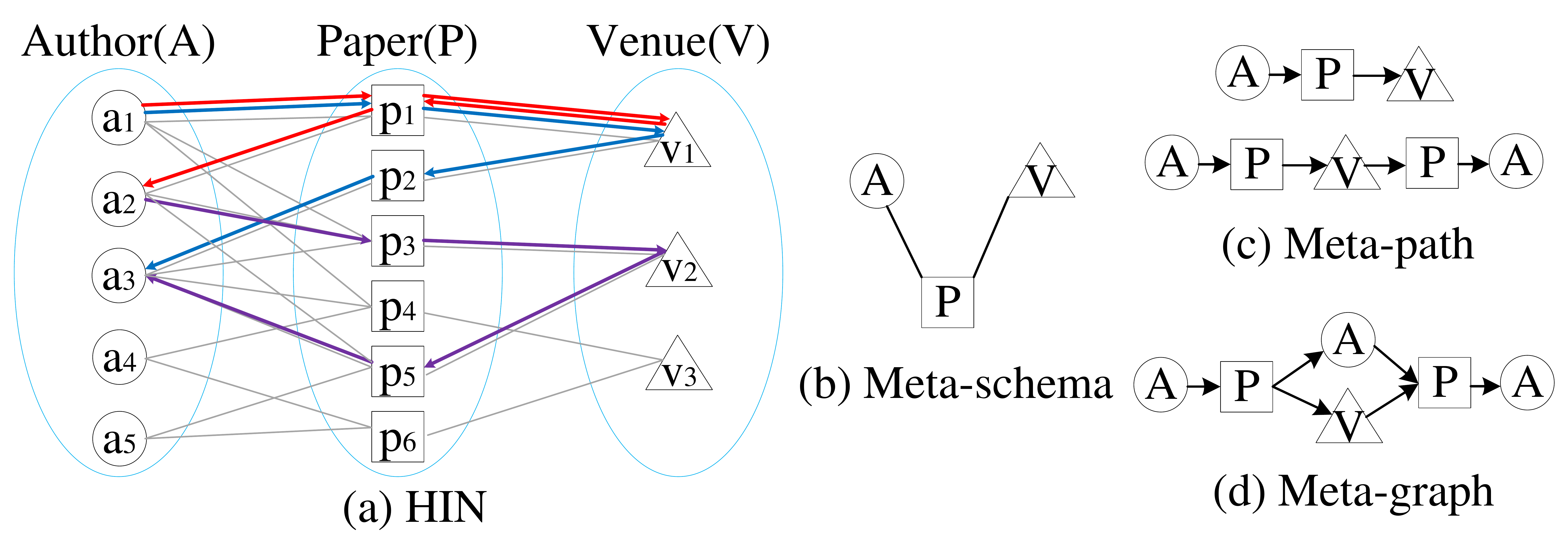}
	\caption{\small An illustrative  heterogeneous scholar network with three entity types: {\it Author}(A), {\it Paper}(P), and {\it Venue}(V), as well as  its meta-schema and some meta-paths/meta-graphs on the meta-schema.}
	\label{Fig:A_HIN}
	\vspace{-0.2cm}
\end{figure}

{Heterogeneous Information Networks} (HINs) have been studied for several years and proven to be useful for many applications~\cite{shi2017survey,cai2018comprehensive}.
The heterogeneous information provided by the types of entities and their relationships in HINs can capture more semantically meaningful information than homogeneous networks~\cite{Sun2011a}.
For example, 
consider a scholar network, as illustrated in Figure~\ref{Fig:A_HIN}, which consists of three entity types: {\it Author}, {\it Paper}, and {\it Venue}, and two relationships: an author {\it writes} a paper, and a paper is {\it published} in a conference venue.
We can introduce two meta-paths~\cite{Sun2011a}: ``{\it Author--Paper--Author} (APA)'' and ``{\it Author--Paper--Venue--Paper--Author} (APVPA)'' which measure the similarity of two authors who co-author many papers or whose papers are published at the same venues. 
The semantics of the two meta-paths give two specific definitions about how the two authors are seen as similar.
Many HIN-specific applications, 
such as entity typing~\cite{RenEWH15,RenEWTVH15}, similarity search~\cite{Sun2011a,SunNHYYY12}, and link prediction~\cite{YuRSGSKNH14,ZhaoYLSL17,HuSZY18}, have been applied to use the semantics of different meta-paths in HINs and have been shown to be useful.

Recently, inspired by the DeepWalk algorithm for homogeneous networks~\cite{PerozziAS14}, meta-path(s) guided random walk based HIN embedding algorithms have also been developed~\cite{YuxiaoDong2017,FuLL17,ChuanShi2017,zhang2018metagraph2vec}.
In general, these algorithms use a two-step approach to generate node embeddings.
First, they perform random walks on an HIN guided by one or mutiple meta-path(s). 
Then they run a Skipgram algorithm which was invented in word embedding approach word2vec~\cite{Mikolov2013} to generate the node embeddings.
The idea of Skipgram algorithm is to use a central node to predict its context nodes given by the random-walk paths.
Despite their success in capturing the heterogeneous information provided by meta-paths and thus learning valuable embedding vectors, 
there are still some problems.
First,
the meta-path guided random walk uses sampling to generate node paths, which are in essence sampled from a higher-order Markov chain.
For example, as illustrated in Figure~\ref{Fig:A_HIN}, for the random walks guided by the  meta-path ``APVPA'', we will constrain that  for the first ``P'' in the meta-path, the next node must be ``V'' and for the second ``P'' in the meta-path, the next node must be ``A''. That is, when we perform next walk from a ``P'', we need to know the previous state before ``P'' being ``A'' or ``V'' to judge the ``P'' being the first or the second. Thus, this is theoretically a second-order Markovian stochastic process.
However, existing algorithms have paid few attention to the essence of the higher-order Markov chain property of meta-path guided random walk, especially to its limiting stationary distribution.
As a random walk can be regarded as a random sampling from the stationary distribution, their inattentive sampling may result in less accurate approximation of the stationary distribution and consequently may have less effective node embeddings.
Second,
compared with single meta-path, meta-graph is able to capture richer information by integrating multiple meta-paths.
However, as different meta-paths characterize different semantics, it is still a challenge how to effectively select and balance multiple meta-paths with appropriate weights.
Moreover, 
the design of meta-paths is usually domain-specific and is difficult for a non-proficient user.
Thus, a more principled way of performing HIN embedding should be fit to extend to a general meta-schema driven embedding model for when proficient design is impracticable.

In this paper, to address the above issues, first,
we  systematically clarify that the meta-path guided random walk is a higher-order Markov chain process.
Then a meta-path based heterogeneous personalized spacey random walk (called {\it HeteSpaceyWalk}) is introduced to improve the walk effectiveness and efficiency by leveraging a non-Markovian spacey strategy.
Given an arbitrary meta-path,
instead of strictly constraining the walks by the meta-path,
the {\it HeteSpaceyWalk} allows spacing out and skipping the intermediate trivial transitions, and in fact performs a special meta-graph guided random walk, which differs from the normal one in four aspects: 
(a) The special meta-graph integrates multiple meta-paths which are automatically produced by folding the original user-given meta-path.
(b) The trade-off of these integrated meta-paths are dynamically self-adjusted. That is, during the walk process, {\it HeteSpaceyWalk} adaptively adjusts the probability of following each possible meta-path according to the carefully designed personalized spacey strategy.
(c) {\it HeteSpaceyWalk} is mathematically guaranteed to attain the same unique limiting stationary distribution with the original meta-path guided higher-order Markovian random walk.
(d) By spacing out and skipping the intermediate trivial transitions, {\it HeteSpaceyWalk} provides a cost-efficient sampling way to be stationary and thus can produce more effective embeddings with less walk-times and walk-length than original meta-path constrained random walk sampling.

Second, by leveraging the {\it HeteSpaceyWalk} process to generate heterogeneous neighborhood, and incorporating them to the Skipgram model, 
we develop a general scalable HIN embedding algorithm: {\it SpaceyMetapath}, which is able to produce semantically meaningful embeddings for multi-typed nodes in an HIN with an arbitrary user-given meta-path.
Further, we extend the guidance from a single meta-path to a user-given meta-graph and a general non-handcrafted meta-schema, and develop two embedding algorithms:  {\it SpaceyMetagraph} and  {\it SpaceyMetaschema}.
The main contributions of our work are summarized as follows:
$\bullet$ Existing approaches have paid few attention to the higher-order Markov chain nature of meta-path guided random walk, especially to its stationary distribution.
We formalize the meta-path guided random walk as a higher-order Markov chain process, and present a heterogeneous personalized spacey random walk to efficiently and effectively attain the expected stationary distribution among nodes, which is based on a solid mathematical foundation~\cite{BensonGL17}.

$\bullet$ We propose a generalized framework for heterogeneous spacey random walk for a meta-path based higher-order Markov chain for HIN embeddings.
Based on this framework, we further extend to an arbitrary meta-graph integrating multiple meta-paths and a general
meta-schema without any handcrafted meta-paths.

$\bullet$ We use extensive experiments in four heterogeneous networks to demonstrate that the proposed methods considerably outperform both conventional homogeneous embedding and heterogeneous meta-path/meta-graph guided embedding methods in two HIN mining tasks, i.e., node classification and link prediction.

The code is available at \url{https://github.com/HKUST-KnowComp/HeteSpaceyWalk}.


\section{Problem Definition}\label{sec:problem-definition}
In this section, we introduce the  formulation of the heterogeneous information network embedding problem.
We first define several key concepts related to HINs as follows~\cite{sun2012mining}.
\vspace{-0.05in}
\begin{definition}\label{def:HIN}
{\it \textbf{Heterogeneous Information Network} (HIN).
An information network is defined as a directed graph ${\mathcal G} = ({\mathcal V}, {\mathcal E})$ with an entity type mapping $\phi$: ${\mathcal V} \to \mathcal A$ and a relation type mapping $\psi$: ${\mathcal E} \to \mathcal R$, where ${\mathcal V}$ denotes the entity set, ${\mathcal E}$ denotes the relation set, $\mathcal A$ denotes the entity type set, and $\mathcal R$ denotes the relation type set. 
When the number of entity types $|\mathcal A|>1$ or the number of relation types $|\mathcal R|>1$, the network is called a heterogeneous information network.
Otherwise, it is called a homogeneous information network.}
\end{definition}
\vspace{-0.05in}
The meta-schema (or network-schema) provides a high-level description of a given heterogeneous information network.
\vspace{-0.05in}
\begin{definition}\label{def:meta_schema}
{\it \textbf{Meta-schema}.
Given an HIN ${\mathcal G} = ({\mathcal V}, {\mathcal E})$ with the entity type mapping $\phi$: ${\mathcal V} \to \mathcal A$ and the relation type mapping $\psi$: $\mathcal E \to \mathcal R$,
the meta-schema (or network-schema) for network $\mathcal G$, denoted as $\mathcal T_{\mathcal G} = (\mathcal A, \mathcal R)$, is a graph with entity types as nodes from $\mathcal A$ and relation types as edges from $\mathcal R$.
}
\end{definition}
\vspace{-0.05in}
Another important concept is the meta-path~\cite{Sun2011a} which defines relationships between entities at the schema level.
\vspace{-0.05in}
\begin{definition}\label{def:meta_path}
{\it \textbf{Meta-path}.
A meta-path $\mathcal P$ is a path defined on the graph of meta-schema $\mathcal{T}_{\mathcal G} = (\mathcal A, \mathcal R)$, with the form of a sequence of node types $\it A_1, A_2,...,A_{L+1}$ and/or relation types $\it R_1, R_2,...,R_L$:
${\it A_1  \xrightarrow{R_1} A_2 \xrightarrow{R_2}  \dots \xrightarrow{R_L} A_{L+1}}$ (or $\mathcal P = {\it (A_1\to A_2\to \ldots \to A_{L+1})}$ when there is no ambiguity),
which defines a composite relation $R = R_1 \cdot R_2 \cdot \ldots \cdot R_L$ between types $A_1$ and $A_{L+1}$, where $\cdot$ denotes relation composition operator, and $L$ is the length of $\mathcal P$.
}
\end{definition}
\vspace{-0.05in}
We call a path $p \!=\! (v_1\!\to\! v_2\!\to\! \ldots\! \to\! v_{L+1})$ between $v_1$ and $v_{L+1}$ in network $\mathcal{G}$ follows the meta-path $\mathcal P$, if $\forall l, \phi(v_l) \!=\! A_l$ and each edge
$e_l \!=\! \langle v_l,v_{l+1} \rangle$ belongs to each relation type $R_l$ in $\mathcal P$.
We call $p$ as a {\it path instance} of $\mathcal P$, denoted as $p \!\in \!\mathcal P$.
Besides the meta-path, 
the meta-graph (or meta-structure) is also very useful
which captures complex semantics by integrating multiple meta-paths
~\cite{FangLZWCL16,HuangZCSML16}.
\vspace{-0.05in}
\begin{definition}\label{def:meta_graph}
	{\it \textbf{Meta-graph}.
		A meta-graph (or meta-structure) is a directed acyclic graph $\mathcal{T}_{\mathcal S} \!=\! (\mathcal A_S, \mathcal R_S)$ defined on the given HIN schema $\mathcal{T}_{\mathcal G} \!= \!(\mathcal A, \mathcal R)$, 
		where ${\mathcal A_S} \!\subseteq \!{\mathcal A}$ and ${ \mathcal R_S} \!\subseteq\! {\mathcal R}$.
		In general, a meta-graph has only a single source entity type $A_s$ (i.e., with 0 in-degree) and a single target entity type $A_t$  (i.e., with 0 out-degree).
		Specially, we call the meta-graph $\mathcal{T}_{\mathcal S} $ with  $A_s\!=\!A_t$ as a {\it \textbf{recursive meta-graph}} because it can be recursively extended by tail-head concatenation.
	}
\end{definition}
\vspace{-0.05in}
As an illustration, Figure~\ref{Fig:A_HIN} shows a heterogeneous network and its meta-schema, as well as the meta-paths and meta-graphs defined at the schema level.
Finally, by considering an HIN as an input, we formally define the problem of HIN embedding as follows.
\vspace{-0.05in}
\begin{definition}
{\it
\textbf{Heterogeneous Information Network Embedding}.
Given a heterogeneous information network, denoted as a graph ${\mathcal G} = ({\mathcal V}, {\mathcal E}; \mathcal A, \mathcal R)$.
Embedding is to learn a function ${\mathnormal f}:{\mathcal V} \to {\mathbb R}^d$ that
projects each node $v \in {\mathcal V}$ to a 
vector in a d-dimensional latent space ${\mathbb R}^d$, $d \ll |{\mathcal V}|$ that are able to capture the structures and semantics of multiple types of nodes and relationships.
}
\end{definition}


\section{Heterogeneous Spacey Random Walk based Embeddings}\label{sec:frameworks}
In this section, we introduce the general HIN embedding frameworks for a meta-path, a meta-graph, and a meta-schema.

\subsection{Meta-path based HIN Embedding}
\label{sec:spaceymetapath}

Given an arbitrary meta-path $\mathcal P \!= \!{\it (A_1\!\to \!A_2\!\to\! \ldots \!\to \!A_{L\!+\!1})}$,  our goal is to learn the semantically meaningful embeddings for all entities under the constraint of $\mathcal P$.
We proceed by extending the meta-path guided random walk based embedding~\cite{YuxiaoDong2017} in view of the scalability for large-scale networks.

\subsubsection{Background of Random Walk over HIN}
Meta-path guided random walk over an HIN was first considered in computing similarities in the path ranking algorithm (PRA)~\cite{LaoC10}.
It defines the following transition probabilities:
\begin{equation}\label{Eq:normalized_adjacency_matrix}
{\bf P}_{\!A_l,A_{l+1}} = {\bf D}_{\!A_l,A_{l+1}}^{-1}{\bf W}_{\!A_l,A_{l+1}},
\end{equation}
where ${\bf W\!}_{\!A_l\!,A_{l+1}}$ is the adjacency matrix between nodes in type $A_l$ and nodes in type $ A_{l+1}$,
and ${\bf D\!}_{\!A_l,A_{l+1}}$ is the degree matrix with  ${\bf D\!}_{\!A_l,A_{l+1}}\!(v_i,\!v_i) \!\!=\! \!\!\sum_j \!\!{\bf W\!}_{\!A_l\!,A_{l+1}}\!(v_i,\!v_j)$.
When performing a random walk from a node $v_i$ in type $A_l$, we choose next node $v_j$ in type $A_{l+1}$ based on the probability ${\bf P}_{\!A_l,A_{l+1}}(\!v_i,\!v_j)$.
In PRA algorithm, it uses the normalized commuting matrix, i.e., ${\bf P\!}_{\!A_1\!,A_2} \!\!\cdot\! {\bf P\!}_{\!A_2\!,A_3}\!\!\cdot\! \ldots \!\cdot \!{\bf P\!}_{\!A_{\!L}A_{\!L+1}}$ to define the meta-path based similarities between nodes of type $A_1$ and $A_{L+1}$,
and claims that this is an ad-hoc definition of hitting/commuting probability. 
The actual expected hitting/commuting probability can be approximated by the stationary distribution of the random walk following the meta-path constraints.
However, as discussed in Section~\ref{sec:introduction},
the existing random walk based HIN embedding methods (e.g.,~\cite{YuxiaoDong2017,FuLL17,ChuanShi2017,zhang2018metagraph2vec}) directly use the random walk sampling but pay few attention to definitely explore the higher-order Markov chain property of meta-path guided random walk, especially to its limiting stationary distribution, which is essential for describing the long-run behavior of the random walk.

\subsubsection{Higher-order Markov Chains}
In this section, 
we first clarify that the meta-path guided random walk is a higher-order Markov chain using the following lemma and its explanations.
\begin{lemma}\label{lemma:meta_path_Markov_property}
{\it
For an arbitrary meta-path $\mathcal P \!=\! {\it (A_1\!\to\! A_2\!\to\! \cdots \!\to\! A_{\!L+\!1})}$, 
we can define a $k$th-order Markov chain, iff $\mathcal P$ can be divided into a set of unique $k$-length meta-paths ${\it (\!A_l\!\to\! A_{l+1}\!\to\! \cdots \!\to \!A_{l+\!k})}$ which satisfy that if the last $k$ states are determined by ${\it (A_l, A_{l+\!1}, \cdots\!, A_{l+\!k-\!1})}$, the current state can only be $A_{l+k}$.
After that, we can obtain the transition probabilities of the $k$th-order Markov chain by concatenating the normalized commuting/adjacency matrix of these $k$-length meta-paths, and such transition probabilities can be used to guide a random walk constrained by $\mathcal P$. 
}
\end{lemma}
Note that to ensure the random walk runs continuously, we generally regard $\mathcal P$ should be cyclic with the same start and end entity types, i.e., $A_1 \!=\! A_{L+1}$ (If not, it is easy to jump back from either start or end to get a symmetric one).
Taking the meta-path 
``APVPA'' illustrated in Figure~\ref{Fig:A_HIN} as an example,
we can divide ``APVPA'' into a set of meta-paths: ``APV'', ``PVP'', ``VPA'', and ``PAP''.
With these factorized meta-paths, walking to the current state is uniquely determined by the last two states.
For example, if and only if the last state is a {\it paper} (P) and the second last state is an {\it author} (A), the current state can be only determined to be a {\it venue} (V).
Thus 
a second-order Markov chain can be used to represent the ``APVPA'' guided random walk.
Further, we can define a hypermatrix (three-dimensional tensor) to denote the transition probability for such second-order Markovian random walk.

\begin{definition}
{\it
A second-order hypermatrix ${\bf H}$ for the meta-path guided second-order Markovian random walk can be defined as:
\begin{equation}\label{Eq:second_order_transition_hypermatrix}
{\bf H}_{i,j,k} \triangleq \left\{ 
\begin{aligned}
& \phi(v_i)=A_l, \\
{\bf P}_{A_{l+1},A_{l+2}}(v_j,v_k),\ \ \  & \phi(v_j)=A_{l+1}\\
& \phi(v_k)=A_{l+2} \\
0,  otherwise &
\end{aligned}
\right.,
\end{equation}
where ${\bf H}_{i,j,k}$ represents the transition probability to node $v_k$, given the last node $v_j$ and the second last node $v_i$;
``$A_lA_{l+1}A_{l+2}$'' is one of  ``APV'', ``PVP'', ``VPA'', or ``PAP''; $\phi(\cdot)$ is the node type mapping; and ${\bf P}_{A_{l+1},A_{l+2}} $ is defined as Eq. (\ref{Eq:normalized_adjacency_matrix}).
}
\end{definition}

Although we have formalized the meta-path guided higher-order Markovian random walk,
there can be some trivial walks.
Due to the nature of higher-order Markov chain, the transitions following a meta-path usually depend on the last several states rather than just the last one. As a result, an entity may play a redundant role in the walks.
For example, we define the meta-path ``APVPA'' based second-order Markovian random walk to extract the author similarity by the semantic ``Two authors published at a venue'', and the walk path instances are illustrated in Figure~\ref{Fig:A_HIN}(a).
Among them, 
we can find the red one $a_1\!\!\to\! \!p_1\!\!\to\!\! v_1\!\!\to\!\! p_1\!\!\to \!\!a_2$ actually expresses the same information as the path instance $a_1\!\!\to \!\!p_1\!\!\to\!\! a_2$ following  ``APA''.
In this case, the intermediate transitions $p_1\!\!\to \!\!v_1\!\!\to\!\! p_1$ are trivial walks and can be skipped, 
i.e., 
when we walked from an author $a_1$ to a paper $p_1$, we then can immediately skip to another author as if the current $p_1$ was just walked from a venue $v_1$.
Such shortened walk path $a_1\!\!\to \!\!p_1\!\!\to\!\! a_2$ can be regarded as a faster instance of meta-path ``APVPA''.
This fact motivates us that instead of strictly constraining the walks by the meta-path, it can be preferred to occasionally skip the intermediate trivial transitions with a principled probability.
Then, the meta-path based higher-order Markovian random walk can be more efficient to attain the expected stationary distribution.
Intuitively, for any meta-path $\mathcal P^* \!\!=\!{\it (A_1\!\to\!\cdots\! A_l\!\to\! \cdots \!A_k\!\cdots \!\to\!\! A_{\!L+\!1})}$ with $A_l\!=\!A_k$, we can probabilistically perform a faster random walk following the folded meta-path ${\it (A_1\!\to\!\cdots\! A_l\!\to \!A_{k+\!1}\!\cdots \!\to\! A_{\!L+\!1})}$, and the walk path can be equivalently regarded as a special instance of meta-path $\mathcal P^*$.
To achieve this, we introduce the following spacey random walk strategy.

\subsubsection{Heterogeneous Personalized Spacey Random Walk}
Given a 
higher-order Markov chain, 
the spacey random walk provides a space-friendly and efficient alternative approximation and is mathematically guaranteed to converge to the same limiting stationary distribution~\cite{li2014limiting,BensonGL17}.
Inspired by that, 
to obtain the embeddings of multi-typed entities in an HIN, we first define our meta-path based heterogeneous personalized spacey random walk (called {\it HeteSpaceyWalk}) as follows.
We use the aforementioned ``APVPA'' based second-order Markov chain and transition hypermatrix ${\bf H}_{i,j,k}$ as an illustration, while higher-order chains are similar.

\begin{definition}\label{def:personalized_spacey_walk}
{\it \textbf{Heterogeneous Personalized Spacey Random Walk for a Meta-path based Second-order Markov Chain}.
Given a second-order Markov chain, the transition hypermatrix ${\bf H}_{i,j,k}$ is concatenated by a set of transition probabilities based on a set of factorized meta-paths (as defined in Eq. (\ref{Eq:second_order_transition_hypermatrix})).
Then, these  transition probabilities can be used to guide a personalized spacey random walk.
Such stochastic process consists of a sequence of states/nodes $X(0), X(1), X(2),\cdots \!,X(n)$, and the probability law is defined to use
\begin{equation}\label{Eq:spacey_out_law}
Pr\!\left\{Y(n)\!=\!v_i | \mathcal{F}_n \right\} \triangleq
\begin{cases}
(1-\alpha) + \alpha{\bf w}_i(n), &v_i=X(n-1) \\
\alpha{\bf w}_i(n), & v_i\not =X(n-1) 
\end{cases}
\end{equation}
to choose the second last state $Y(n)$, and then use
\begin{equation}\label{Eq:spacey_walk}
Pr\!\left\{X(n+1)=v_k | X(n)=v_j, Y(n)=v_i \right\}  \triangleq  {\bf H}_{i,j,k}
\end{equation}
to choose the next node, where $\mathcal{F}_n$ is the $\sigma$-field generated by the random variables $X(i), i\!=\!1,2,...,n$, $X(0)$ is the initial state; 
$\alpha \in (0,1)$  is a hyper-parameter that can control a user's personalized behavior;
${\bf w}(n)$ is the occupation vector at step $n$ and is defined as:
\begin{equation}\label{Eq:occupation_vector}
{\bf w}_i(n) \triangleq \frac{1}{n+ N}\left( 1+\sum_{s=1}^{n} Ind\left\{X(s)=v_i\right\}\right),
\end{equation}
where $N$ is the number of total states.
}
\end{definition}
In short, 
with personalization, once the spacey random walker visits $X(n)$ at step $n$, it spaces out and forgets its second last state (i.e., the state $X(n-1)$) with probability $\alpha$.
It then invents a new history state $Y(n)$ by randomly drawing a past state $X(1),...,X(n)$.
Then it transitions to $X(n+1)$ as if its last two states were $X(n)$ and $Y(n)$.
Without personalization (i.e. $\alpha\!=\!0$), the spacey random walker performs a normal second-order Markovian random walk.

\begin{figure*}
	\small
	\vspace{-0.1in}
	\setlength{\abovecaptionskip}{-0.1cm}
	\setlength{\belowcaptionskip}{-0.4cm}
	\centering
	\includegraphics[width=0.9\textwidth]{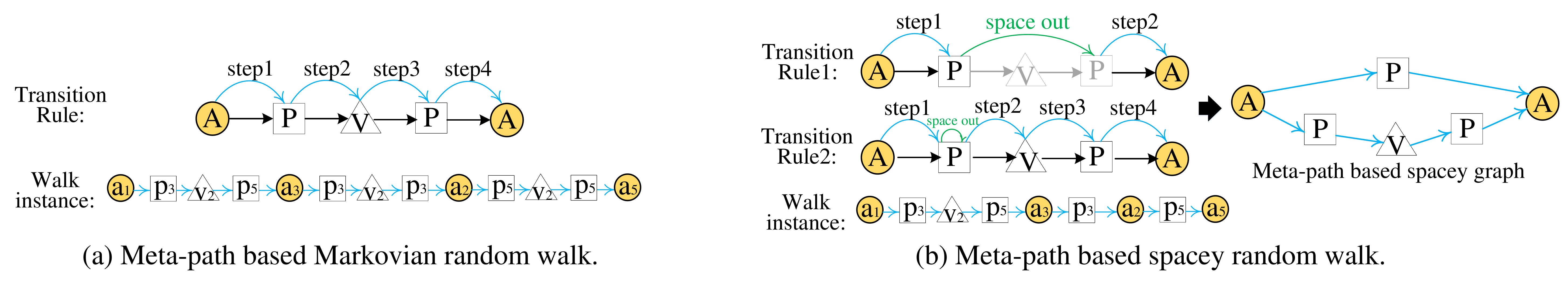}
	\caption{\small An intuitive comparison of meta-path based Markovian random walk and spacey random walk.
		When the spacey random walker walks from ``P'' to the next,  it forgets its second last state is ``A'' and uses Eq.~(\ref{Eq:spacey_out_law}) to choose a new state which is assumed as the precursor of ``P''.
		Constrained by ``APVPA'' based second-order Markov chain, the assumed precursor of ``P'' can be ``V'' or ``A'', and respectively, there are two rules to guide the next transition: directly walk to ``A'' ({\it rule1}), or walk to ``V'' as a normal Markovian random walk ({\it rule2}).
	}
	\label{Fig:comparison_random_walk}
\end{figure*}

Figure~\ref{Fig:comparison_random_walk} shows an intuitive example of comparing normal meta-path based Markovian random walk and our proposed meta-path based spacey random walk, given the meta-path ``APVPA'' on the scholar network illustrated in Figure~\ref{Fig:A_HIN}(a).
We can see that, different from the Markovian random walk which strictly follows the constraints of the user-given meta-path,
the spacey random walk allows skipping the intermediate transitions to improve walk efficiency and quality, which is actually a shortened random walk following the folded sub-path of the original user-given meta-path.
In fact, as illustrated in Figure~\ref{Fig:comparison_random_walk}(b),
the spacey strategy produces a special meta-graph/multi-meta-paths (which we call as the {\it meta-path based spacey graph} integrating the original user given meta-path and its folded sub-meta-paths) guided random walk process.
As a result, the proposed spacey random walk can efficiently and effectively capture richer information and thus achieve better performance than the normal Markovian random walk.
For example, we can see the two walk instances in Figure~\ref{Fig:comparison_random_walk}, which indicate that the spacey random walk can use shorter walk steps to capture richer relationships than normal Markovian random walk.
Moreover,
different from directly deploying meta-graph/multi-meta-paths guided random walk, which is problematic to determine appropriate proportions to balance multiple meta-paths,
the proposed spacey random walk adaptively adjusts the probability of following  the original meta-path or any folded sub-meta-path  according to  a carefully designed personalized occupation probability (as defined in Eq.~(\ref{Eq:spacey_out_law})) calculated dynamically by past states.
The following theorem shows the guarantee of
only more efficiently and effectively extracting the specific semantic of domain-proficient meta-path provided by users, i.e., the proposed spacey strategy will not absorb other heterogeneous information beyond the users' concern.
\begin{theorem}\label{theorem-convergence}
The limiting distribution of the heterogeneous personalized spacey random walk for a meta-path equals to the same stationary distribution of the original meta-path based higher-order Markovian random walk under the rank-one approximation condition.
\end{theorem}


The proof follows the spacey random walk theory~\cite{BensonGL17} and  $z$ eigenvector solution theory~\cite{li2014limiting} which point out that by considering a ``rank-one approximation'' for a transition hypermatrix ${\bf H}$ of second-order Markov chain, the stationary distribution formula reduces to
\begin{equation}
{\bf \pi}_k = \sum_{i,j}{\bf H}_{i,j,k}{\bf \pi}_j {\bf \pi}_i, \ \ \sum_k {\bf \pi}_k =1, {\bf \pi}_k \ge0, N\ge k\ge 1,
\end{equation}
where ${\bf \pi}_k$ is node distribution and $N$ is the number of nodes.
Following this,
we next demonstrate that the unique stationary solution of meta-path based second-order Markov chain is also the limiting distribution of the proposed personalized spacey random walk.

First, we can formally express that the defined spacey random walker $\{X(n)\}$ approximates a second-order Markov chain with a first-order transition probability matrix:
\begin{equation}\label{Eq:probabilities_matrix_spacey_walk}
\begin{split}
{\bf R}_{j,k} & \triangleq \sum_i {\bf H}_{i,j,k}\left( (1-\alpha){\bf x}_i(n)+\alpha{\bf w}_i(n) \right) \\
& = [{\bf H}\cdot(((1-\alpha){\bf x}(n) + \alpha{\bf w}(n))\otimes {\bf I})]_{j,k},
\end{split}
\end{equation}
where ${\bf x}(n)$ is the nodes distribution at step $n$, ${\bf w}(n)$ is the occupation probability vector at step $n$ (as defined in Eq.~(\ref{Eq:occupation_vector})).

Then,
if the process  $\{X(n)\}$  converges to a unique stationary distribution ${\bf x}(n)$ satisfying ${\bf R}{\bf x}(n)={\bf x}(n)$,
for the case  $n\gg L > 0$, we have:
\begin{equation*}
{\bf w}(n+L) \thickapprox \frac{n{\bf w}(n)+L{\bf x}(n)}{n+L}={\bf w}(n)+\frac{L}{n+L}({\bf x}(n)-{\bf w}(n)).
\end{equation*}
In a continuous time limit $L\to 0$, we have:
\begin{equation*}
\frac{d{\bf w}(n)}{dL}\thickapprox \lim_{L\to 0} \frac{{\bf w}(n+L)-{\bf w}(n)}{L}=\frac{1}{n}({\bf x}(n)-{\bf w}(n)).
\end{equation*}
Thus,  if this process converges, it must converge to a point where ${\bf x}(n)={\bf w}(n)$.
Further, we can find that the limiting distribution ${\bf x}$ heuristically satisfy:
\begin{equation*}
{\bf x}={\bf R}{\bf x}={\bf H}\cdot({((1-\alpha){\bf x}+ \alpha{\bf w})}\otimes {\bf I}){\bf x}={\bf H}\cdot({\bf x}\otimes {\bf I}){\bf x}={\bf H}\cdot({\bf x}\otimes {\bf x}).
\end{equation*}
That is,
\begin{equation*}
{\bf x}_k=\sum_{i,j}{\bf H}_{i,j,k}{\bf x}_j {\bf x}_i.
\end{equation*}

Therefore, the limiting distribution of matrix $ {\bf R}$  is also the unique stationary solution of hypermatrix ${\bf H}$.
The defined heterogeneous personalized spacey random walk provides an efficient alternative approximation for the meta-path based higher-order Markovian random walk without changing the limiting stationarity.

\subsubsection{Spacey Random Walk based Embedding}
Now we can use the heterogeneous personalized spacey random walk to generate node sequences in an HIN, and then feed them to the Skipgram model~\cite{mikolov2013efficient,Mikolov2013} to learn node embeddings.
We follow~\cite{YuxiaoDong2017} to define the heterogeneous Skipgram model.
Given the generated paths corpus $\mathcal{V_P}$ guided by meta-path $\mathcal P$, the model is to minimize following objective function:
\begin{equation}\label{Eq:skip-gram}
  \mathop{\arg\min}_{\theta} \ -\sum_{v_i\in \mathcal{V_P}}\sum_{A\in \mathcal{A}}\sum_{v_j \in {\mathcal N}_A(v_i)}\log Pr(v_j|v_i,\theta),
\end{equation}
where ${\mathcal N}_A(v_i)$ denotes $v$'s neighborhood with type $A$. For each pair of entities $(v_i,v_j)$, their joint probability $Pr(v_j|v_i,\theta)$ is commonly defined as such softmax function:
\begin{equation}\label{Eq:softmax}
Pr(v_j|v_i,\theta) = \frac{\exp\{{\bf u}_j^T {\bf v}_i\}}{\sum_{j'}\exp\{{\bf u}_{j'}^T {\bf v}_i\}},
\end{equation}
where ${\bf u}_j$ is the context vector of $v_j$ and ${\bf v}_i$ is the embedding vector of $v_i$.
We also use the negative sampling technique~\cite{Mikolov2013} for optimizing Eq.~(\ref{Eq:skip-gram}) which is modified as: \begin{equation}\log \sigma({\bf u}_j^T {\bf v}_i\}) +
  \sum_1^m \mathbb{E}_{v_c \thicksim P_n(v_i)} \left[\log \sigma(-{\bf u}_c^T {\bf v}_i\}) \right],
\end{equation}
where $\sigma(x)\!=\!\frac{1}{1+\exp\{-x\}}$ and $P_n(v_i)$ is the 
sampling distribution.


We call the above approach as {\it SpaceyMetapath} algorithm.

\subsection{Meta-graph based HIN Embedding}
\label{sec:spaceymetagraph}

In this section, 
we further propose a meta-graph based spacey random walk algorithm, named  {\it SpaceyMetagraph}.
Compared with single meta-path, a meta-graph can capture more complex heterogeneous structures and richer semantics by integrating multiple meta-paths.
The challenge is how to concatenate these different higher-order Markov chains with appropriate proportions, i.e., how to balance the branch 
choices in a meta-graph.
For example, for the meta-graph in Figure~\ref{Fig:A_HIN}(d), 
when we need to perform random walk from the first ``P" to its successors, there are two choices: ``A" or ``V''.
Unlike a higher-order Markovian transition, this transition does not depend on any previous state.
%
To deal with this issue, we extend our spacey random walk principle from choosing previous states to choosing the branch choices,
and define our meta-graph based heterogeneous personalized spacey random walk as follows.

\begin{definition}\label{def:personalized_spacey_walk_metagraph}
	{\it \textbf{Meta-graph based Heterogeneous Personalized Spacey Random Walk}.
		Given $M$ second-order Markov chains based on a meta-graph $\mathcal{T}_{\mathcal S} = (\mathcal A_S, \mathcal R_S)$ integrating $M$ meta-paths,
		an integrated transition hypermatrix ${\bf \overline{H}}$ for this meta-graph is defined as:
		\begin{equation}
		{\bf \overline{H}}_{i,j,k}=\frac{1}{\sum_{m} Ind \{{\bf H}_{i,j,k}^m >0\}}\sum_{m} {\bf H}_{i,j,k}^m,
		\end{equation}
		where ${\bf H}^m$ is the transition probabilities of $m$-th Markov chain as defined in Eq.~(\ref{Eq:second_order_transition_hypermatrix}).
		Then, the probability law to guide the meta-graph based heterogeneous personalized spacey random walk 
		is defined to use the personalized occupation probability (as defined in Eq.~(\ref{Eq:spacey_out_law})) to choose the second last state $Y(n)$, 
		and use
		\begin{equation}\label{Eq:spacey_type}
		Pr\!\{\Phi \!(n\!+\!1)\!\!=\!\!A | X\!(n)\!\!=\!\!v_j,  \! Y\!(n)\!\!=\!\!v_i \}\! \! \triangleq \!\!
		\begin{cases}
		\!(1\!\!-\!\!\alpha)\frac{1}{|\!N_{\!{\small \mathcal{T}_{\!\mathcal S}}}\!(v_i\!,\!v_j)\!|}\!+\!\alpha {\bf z}_{\!A}\!(n),\!\!&\! \!A\!\!\in \!\! N_{\!\mathcal{T}_{\!\mathcal S}}\!(v_i\!,\!v_j) \\
		0, & otherwise
		\end{cases}
		\end{equation}
		to choose the next node type $\Phi(n+1)$, and then use
		\begin{equation}\label{Eq:spacey_walk_metagraph}
		Pr\!\left\{X\!(n\!\!+\!\!1)\!=\!v_k | X\!(n)\!=\!v_j, Y\!(n)\!=\!v_i, \Phi(n\!\!+\!\!1)\!= \!A \right\}  \!\!\triangleq \!\!
		\begin{cases}
		{\bf \overline{H}}_{i,j,k},\! & \!\phi(v_k)\!=\!A \\
		0, & \phi(v_k) \! \not =\!A
		\end{cases}
		\end{equation}
		to choose the next node, where $N_{\!{\small \mathcal{T}_{\!\mathcal S}}}\!(v_i,\!v_j)$ is the set of successor types of edge $(\phi (v_i),\phi (v_j))$
		 in $\mathcal{T}_{\!\mathcal S}$, $\phi(\cdot)$ is the node type mapping; 
		$ {\bf z}(n)$ is the partial occupation vector at step $n$:
		\begin{equation}\label{Eq:occupation_vector_metagraph}
		{\bf z}_{\!A}(n) \triangleq \frac{\sum_{\phi (v_i)=A } {\bf w}_i(n)}{\sum_{\phi (v_i)\in N_{\!{\small \mathcal{T}_{\!\mathcal S}}}\!(X\!(n),\!Y\!(n))} {\bf w}_i(n)},\ \  A\!\!\in \!\! N_{\!\mathcal{T}_{\!\mathcal S}}\!(X\!(n),\!Y\!(n)).
		\end{equation}
	}
\end{definition}
In short, with personalization, when the meta-graph based spacey random walker visits $X(n)$ at step $n$, it first forgets its second last state (i.e., the state $X(n-1)$) with a probability $\alpha$ and invents a new history state $Y(n)$ by randomly drawing a past state $X(1),...,X(n)$.
Next, if there are 
branched node types,
it chooses the next type $\Phi(n+1)$  according to their history distribution with probability $\alpha$.
Then it transitions to $X(n+1)$ as if its last two states were $X(n)$ and $Y(n)$ and current node type $\phi (X(n+1))$ is $\Phi(n+1)$.
Without personalization, it performs a non-spacey random walk and randomly chooses the next one from branched node types.

We develop the {\it SpaceyMetagraph} algorithm by leveraging the meta-graph based personalized spacey random walk to generate heterogeneous neighborhood, and then incorporating them to the Skipgram model to produce effective HIN embeddings.

\subsection{Meta-schema based HIN Embedding}
\label{sec:spaceymetaschema}
Although the meta-path and meta-graph have proved to be useful to capture heterogeneous semantics in an HIN,
it is quite difficult for a non-proficient user to design appropriate meta-paths or meta-graphs.
Therefore, it is considerable to extend meta-path/meta-graph  driven  embedding algorithms to a general meta-schema driven embedding algorithm.
Unlike a meta-path or a meta-graph, the meta-schema does not contain duplicate types. 
Thus, the meta-schema guided random walk  is a first-order Markov chain and we do not need to space out  to choose previous states.
However, similar to the meta-graph guided random walk, there is a balance issue for branch choices in a meta-schema.
Therefore, we follow the {\it SpaceyMetagraph} algorithm to 
formalize the meta-schema based heterogeneous personalized random walk as follows.

\begin{definition}\label{def:personalized_spacey_walk_metaschema}
	{\it \textbf{Meta-schema based Heterogeneous Personalized Random Walk}.
		Given an HIN ${\mathcal G} = ({\mathcal V}, {\mathcal E})$ and its meta-schema $\mathcal T_{\mathcal G} = (\mathcal A, \mathcal R)$, 
		the transition probability matrix ${\bf P}_{A_l,A_k}$ starting from type $A_l$ to type $A_k$ is defined
similar as Eq.~(\ref{Eq:normalized_adjacency_matrix}).
		Then, the probability law to guide the meta-schema based heterogeneous personalized random walk
		is defined to use
		\begin{equation}\label{Eq:spacey_type_metaschema}
		Pr\!\{\Phi \!(n\!+\!1)\!\!=\!\!A | X\!(n)\!\!=\!\!v_i \}\! \! \triangleq \!\!
		\begin{cases}
		\!(1\!\!-\!\!\alpha)\frac{1}{|N_{\!{\small \mathcal{T}_{\!\mathcal G}}}\!(v_i\!)|}\!+\!\alpha {\bf z}_{\!A}\!(n),\!\!&\! \!A\!\!\in \!\! N_{\!\mathcal{T}_{\!\mathcal G}}\!(v_i) \\
		0, & otherwise
		\end{cases}
		\end{equation}
		to choose the next node type $\Phi(n+1)$, and then use
		\begin{equation}\label{Eq:spacey_walk_metaschema}
		Pr\!\left\{X\!(n\!\!+\!\!1)\!=\!v_j | X\!(n)\!=\!v_i, \!\Phi(n\!\!+\!\!1)\!= \!\!A \right\}  \!\!\triangleq \!\! 
		\begin{cases}
		{\bf P}_{\!\phi\!(\!v_i\!),\!A}(v_i,\!v_j),\!\! & \!\!\phi(v_j)\!\!=\!\!A \\
		0, \!\!&\!\! \phi(v_j) \!\! \not =\!\!A
		\end{cases}
		\end{equation}
		to choose the next node, where $N_{\!{\small \mathcal{T}_{\!\mathcal G}}}\!(v_i)$ is the set of adjacency types of $\phi (v_i)$
		in $\mathcal{T}_{\!\mathcal G}$; 
		$ {\bf z}(n)$ is the partial occupation vector at step $n$:
		\begin{equation}\label{Eq:occupation_vector_metaschema}
		{\bf z}_{\!A}(n) \triangleq \frac{\sum_{\phi (v_i)=A } {\bf w}_i(n)}{\sum_{\phi (v_i)\in N_{\!{\small \mathcal{T}_{\!\mathcal G}}}\!(X\!(n))} {\bf w}_i(n)},\ \  A\!\!\in \!\! N_{\!\mathcal{T}_{\!\mathcal G}}\!(X\!(n)).
		\end{equation}
	}
\end{definition}

Note that the meta-schema based heterogeneous personalized spacey random walk is a special case of the meta-graph based one where the meta-graph only has first-order Markov chains.
It is different from directly performing random walk on the HIN by treating it as a homogeneous graph, as it dynamically determines what is the type of next node and choose a random node under the chosen type.
Given the random walk based node sequences, the Skipgram model can be used to  learn  embeddings as described in Section~\ref{sec:spaceymetapath}.
We similarly denote this general framework as  {\it SpaceyMetaschema}.

\begin{figure*}
	\vspace{-0.2in}
	\setlength{\abovecaptionskip}{-0.1cm}
	\setlength{\belowcaptionskip}{-0.2cm}
	\centering
	\subfigure[\small ACM]{
		\label{Fig:ACM_dataset}
		\includegraphics[width=0.23\textwidth]{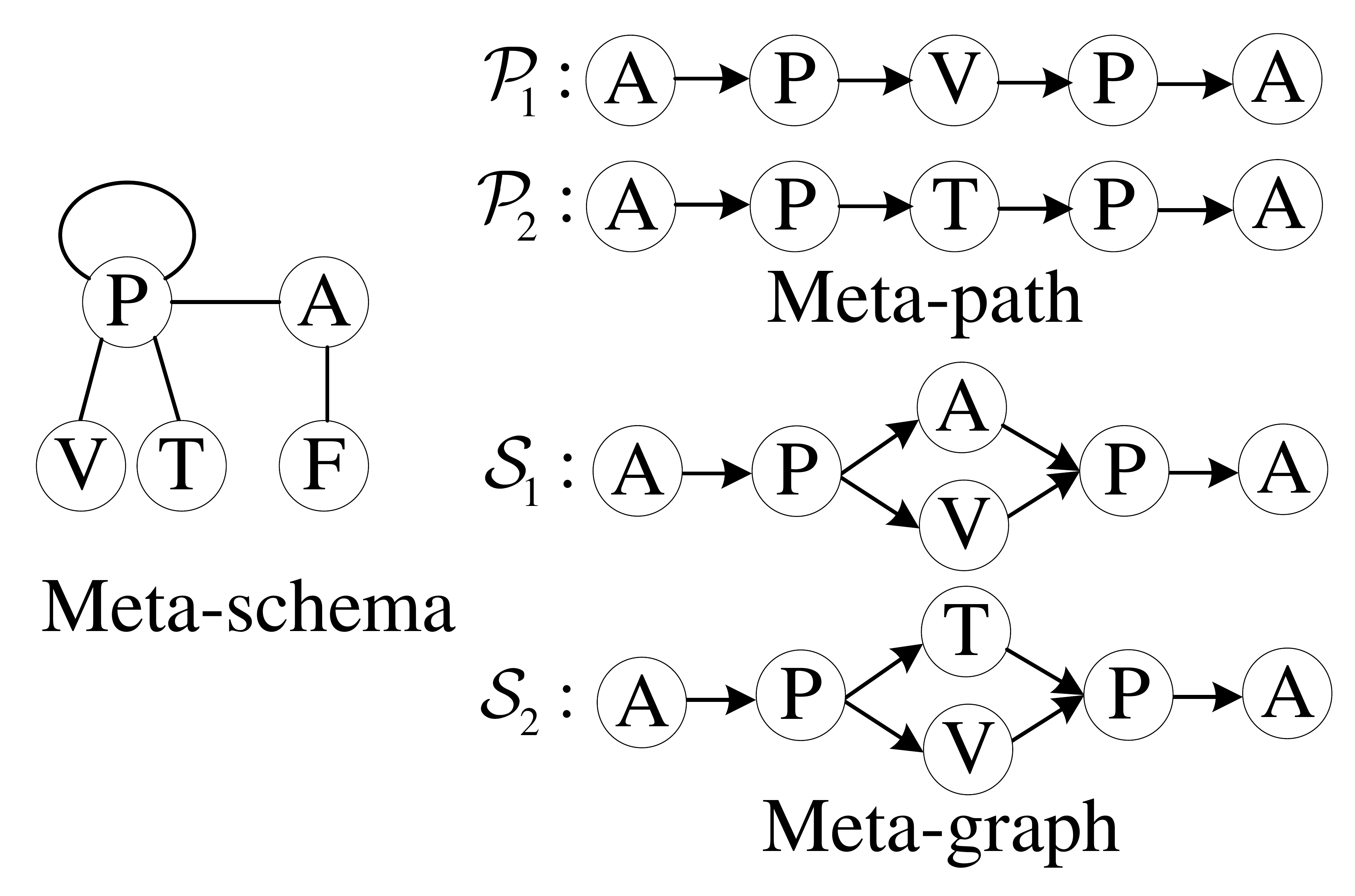}
	}
	\subfigure[\small DBLP]{
		\label{Fig:DBLP_dataset}
		\includegraphics[width=0.23\textwidth]{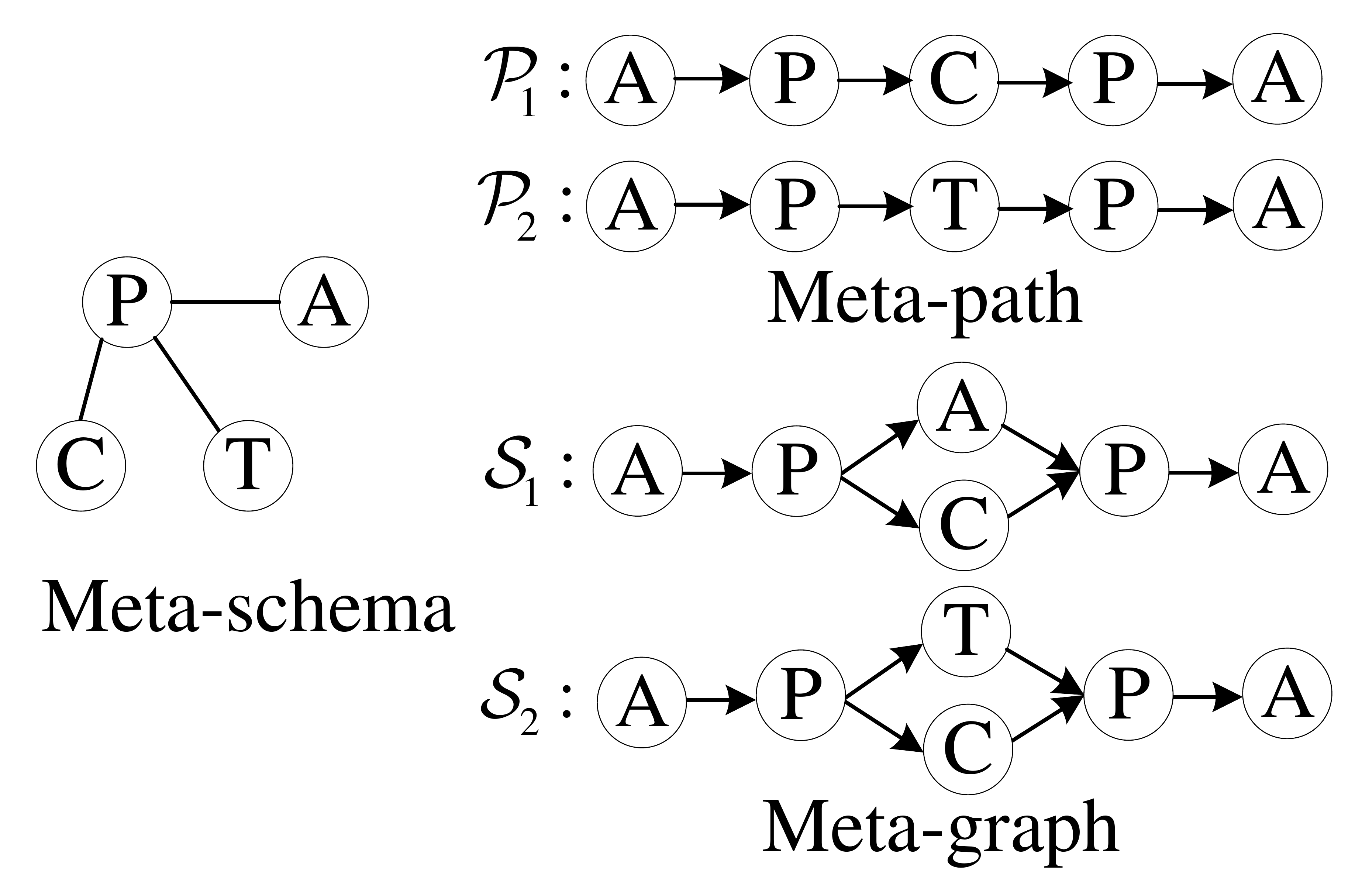}
	}
	\subfigure[\small Douban]
	{\label{Fig:Douban_dataset}
		\includegraphics[width=0.24\textwidth]{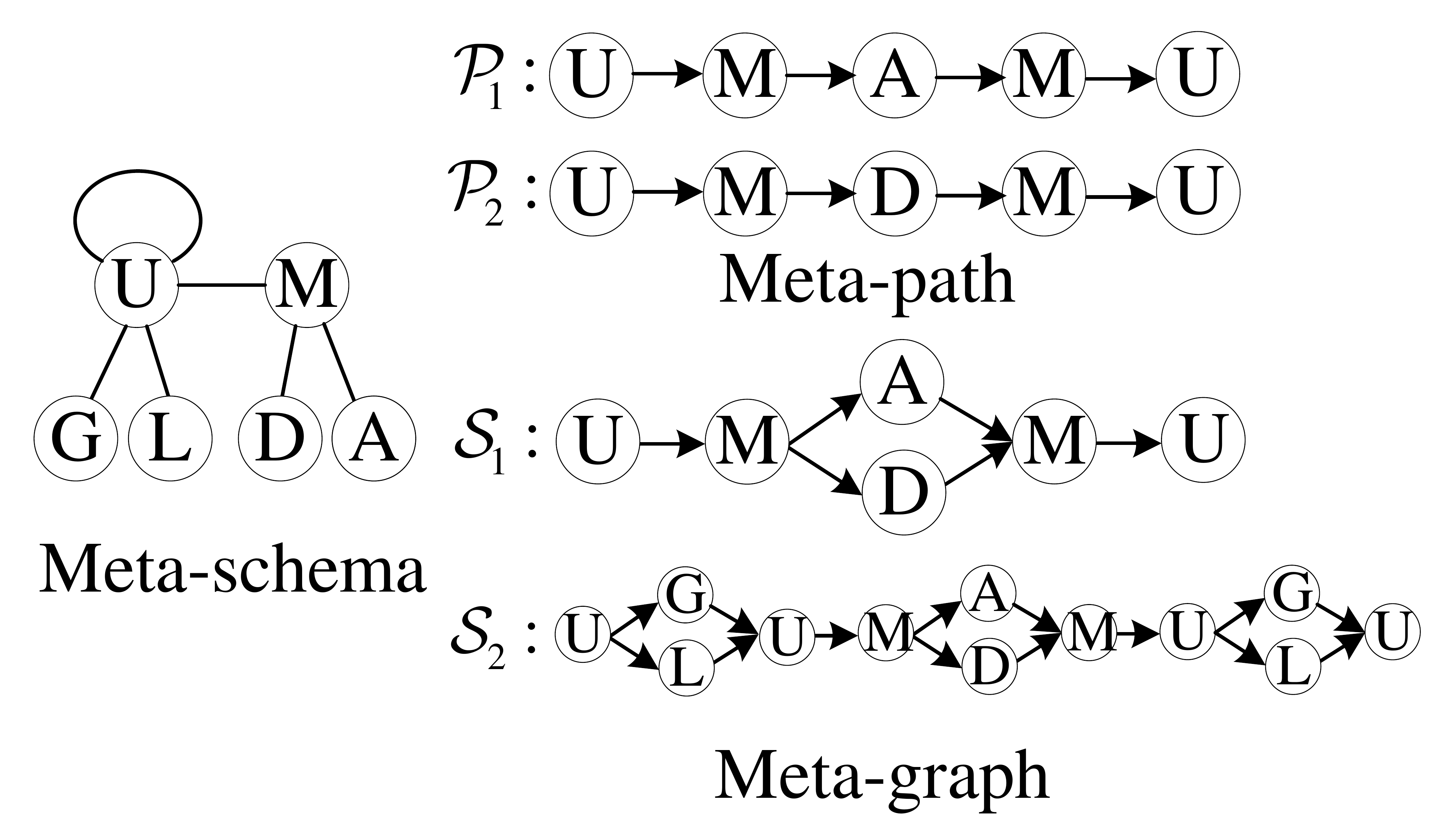}
	}
	\subfigure[\small Yelp]
	{\label{Fig:YELP_dataset}
		\includegraphics[width=0.23\textwidth]{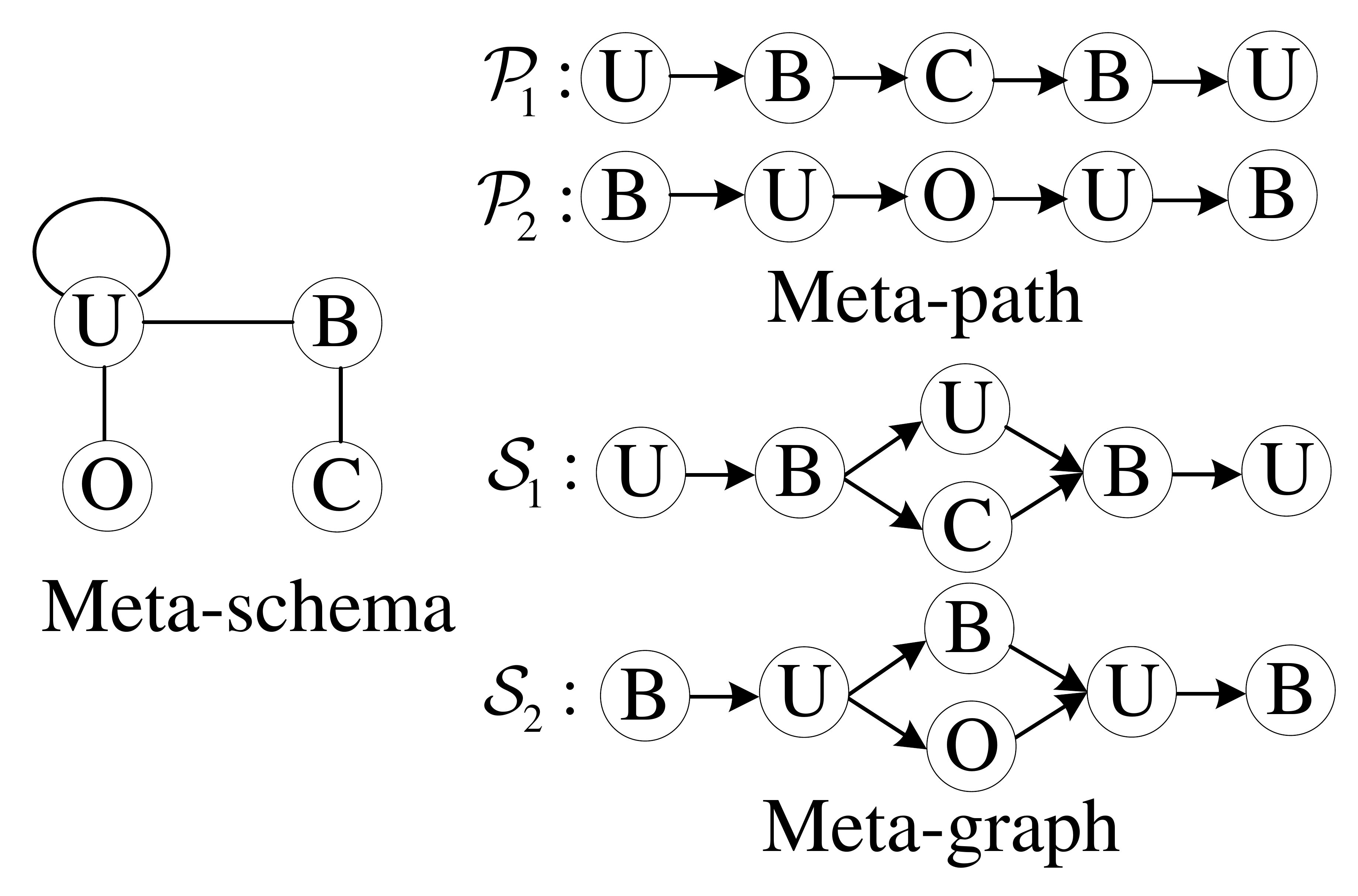}
	}
	\caption{\small The meta-schema, meta-paths, and meta-graphs of each dataset used in the experiments.}
	\label{Fig:datasets}
	\vspace{-0.1in}
\end{figure*}

\section{Experiments}\label{sec:experiments}
In this section, we show the experimental results.
\vspace{-0.1in}
\subsection{Datasets}
\label{sec:dataset}
Here we use the following four heterogeneous networks:




 $\bullet$ \textbf{ACM:}
 The ACM dataset~\cite{shi2014hetesim} 
 contains 196 venues (V), 12,499 papers (P), 1,533 terms (T), 
 17,431 authors (A), and 1,804 author affiliations (F).
The  meta-schema of  ACM  dataset is shown  in Figure~\ref{Fig:ACM_dataset}.

 $\bullet$ \textbf{DBLP:}
The dataset contains 14,376 papers (P), 20 conferences (C), 14,475 authors (A), and 8,920 terms (T).
The meta-schema of DBLP dataset is shown in Figure~\ref{Fig:DBLP_dataset}.

 $\bullet$ \textbf{Douban:}
The Douban dataset~\cite{shi2015semantic} 
 contains 13,367 users (U), 12,677 movies (M), 2,753 groups (G), 349 locations (L), 2,449 directors (D), and 6,311 actors (A).
The meta-schema of Douban dataset is shown in Figure~\ref{Fig:Douban_dataset}.
 
  $\bullet$ \textbf{Yelp:}
  The Yelp dataset~\cite{shi2015semantic} 
 contains 16,239 users (U), 14,284 businesses (B), 47 cities (C), and 11 compliments (O).
The meta-schema of Yelp dataset is shown in Figure~\ref{Fig:YELP_dataset}.




\vspace{-0.2cm}
\subsection{Experimental Settings}
\label{sec:exp_settings}
We compare the following state-of-the-art homogeneous and heterogeneous network embedding methods.


$\bullet$ \textbf{DeepWalk}~\cite{PerozziAS14} is a recently proposed homogeneous network embedding model, which learns $d$-dimensional node vectors by capturing the contextual information via uniform random walks.

$\bullet$ \textbf{LINE}~\cite{TangQWZYM15} 
is a method that preserves first-order and second-order proximities between nodes separately.
We use the suggested version to learn two $d/2$-dimensional vectors (one for each-order) and then concatenate them.


$\bullet$ \textbf{PTE}~\cite{Tang2015PTE} 
is an extension of LINE for heterogeneous network embedding, which decomposes an HIN to a set of bipartite networks by edge types.


$\bullet$ \textbf{Metapath2vec}~\cite{YuxiaoDong2017} is the current state-of-the-art network embedding method for HINs, which formalizes meta-path-based random walks to generate heterogeneous neighborhood 
and then leverages a heterogeneous skipgram model to perform node embeddings.

$\bullet$ \textbf{Metagraph2vec}~\cite{zhang2018metagraph2vec} uses a meta-graph to guide the generation of random walks and to learn latent embeddings for multiple types of nodes in HINs.

We evaluate the quality of embedding vectors learned by different methods over two classical heterogeneous network mining tasks:
multi-label node classification and link prediction. 
For the common hyperparameters, we set
learning rate $lr\!=\!0.025$,  
negative samples $m\!=\!5$, and the embedding dimension $d\!=\!128$ for a trade-off between the computational time and accuracy.
For random walk based methods, 
we set walk times per node $t\!=\!20$, walk length of each walk $l\!=\!320$, neighborhood size $w\!=\!10$.
Specially, for our proposed {\it SpaceyMetapath}, {\it SpaceyMetagraph}, and {\it SpaceyMetaschema}, we set the personalized probability $\alpha\!=\!0.8$.
For LINE and PTE, we set the total number of samples as 100 million.
For meta-path driven (Metapath2vec, SpaceyMetapath) and meta-graph driven (Metagraph2vec, SpaceyMetagraph) methods,
we survey most of the meta-path/meta-graph based work and empirically use some meta-paths and meta-graphs (which has been proven to be the most commonly effectively used schemes to extract heterogeneous semantics~\cite{Sun2011a,shi2014hetesim,YuxiaoDong2017,zheng2017recommendation,zhang2018metagraph2vec}).
The selected meta-paths and meta-graphs of each dataset 
are shown in Figure~\ref{Fig:datasets}.

\begin{table*}[!htb]
 	\small
	\caption{\small The Micro-F1 scores for multi-label node classification. ``--'' indicates that the used meta-path/meta-graph cannot generate embeddings for nodes of target type.}\label{Tab:classify_results}
	\vspace{-0.15in}
	\centering
	 \renewcommand\arraystretch{0.9}
	\begin{tabular}{p{2.8cm}<{\centering}|p{0.9cm}<{\centering}|p{0.9cm}<{\centering}|p{0.9cm}<{\centering}|p{0.9cm}<{\centering}|p{0.9cm}<{\centering}|p{0.9cm}<{\centering}|p{0.9cm}<{\centering}|p{0.9cm}<{\centering}|p{0.9cm}<{\centering}|p{0.9cm}<{\centering}|p{0.9cm}<{\centering}}
		\toprule
		Dataset & \multicolumn{3}{c|}{ACM} & \multicolumn{2}{c|}{DBLP} & \multicolumn{4}{c|}{Douban} & \multicolumn{2}{c}{Yelp} \\
		\midrule
		Node Type & Paper & Author & Venue & Paper & Author & User & Movie & Director & Actor & User & Business \\
		\midrule
		DeepWalk &  
0.5746 & 0.5560 & 0.8198 & 0.6590 & 0.9231 & 0.7205 & 0.5606 & 0.6370 & 0.6592 & 0.5457 & 0.3393 \\
LINE &
0.5608 & 0.5887 & 0.8169 & 0.7120 & 0.9379 & 0.7431 & 0.5621 & 0.6396 & 0.6511 & 0.5324 & 0.3467 \\ 
PTE &
0.5613 & 0.5903 & 0.6838 & 0.6680 & 0.9368 & 0.7077 & 0.5595 & 0.5674 & 0.6260 & 0.5178 & 0.3360 \\
\midrule
Metapath2vec-$\mathcal P_{1}$ & 
0.5774 & 0.6004 & 0.7826 & 0.8220 & 0.9397 & 0.7681 & 0.5678 & -- & 0.6661 & 0.5585 & 0.3435 \\
Metapath2vec-$\mathcal P_{2}$ & 
0.5791 & 0.6164 & -- & 0.5920 & 0.8627 & 0.7622 & 0.5454 & 0.6225 & -- & 0.5648 & 0.2864 \\ 
\hline
Metagraph2vec-$\mathcal S_{1}$ & 
0.5688 & 0.6122 & 0.7880 & 0.8273 & 0.9365 & 0.7924 & 0.5842 & 0.6509 & 0.6669 & 0.6045 & 0.3529 \\ 
Metagraph2vec-$\mathcal S_{2}$ & 
0.6108 & 0.6486 & 0.7990 & 0.8076 & 0.9506 & 0.7953 & 0.5535 & 0.6115 & 0.6366 & 0.6012 & 0.3482 \\ 
\midrule
SpaceyMetapath-$\mathcal P_{1}$ &
0.5805 & 0.6140 & 0.8185 & 0.8410 & 0.9439 & 0.7864 & 0.5711 & -- & 0.6839 & 0.6006 & 0.3574 \\
SpaceyMetapath-$\mathcal P_{2}$ & 
0.5880 & 0.6413 & -- & 0.6300 & 0.8836 & 0.7768 & 0.5596 & 0.6368 & -- & 0.5870 & 0.3312 \\
\hline
SpaceyMetagraph-$\mathcal S_{1}$ & 
0.5750 & 0.6171 & 0.8094 & {\bf 0.8520} & 0.9438 & {\bf 0.8036} & {\bf 0.5889} & {\bf 0.6676} & {\bf 0.6906} & {\bf 0.6101} & {\bf 0.3627} \\
SpaceyMetagraph-$\mathcal S_{2}$ & 
0.6136 & {\bf 0.6574} & {\bf 0.8412} & 0.8410 & {\bf 0.9521} & 0.8017 & 0.5720 & 0.6522 & 0.6809 & 0.6072 & 0.3524 \\ 
\hline
SpaceyMetaschema & 
{\bf 0.6145} & 0.6493 & 0.8369 & 0.8330 & 0.9512 & 0.8032 & 0.5793 & 0.6535 & 0.6832 & 0.5688 & 0.3536 \\ 
		\bottomrule
	\end{tabular}
	 \vspace{-0.05in}
\end{table*}

\begin{table*}[!htb]
	\small
	\caption{\small The Macro-F1 scores  for multi-label node classification. ``--'' indicates that the used meta-path/meta-graph cannot generate embeddings for nodes of target type.}\label{Tab:classify_results_macro}
	\vspace{-0.15in}
	\centering
	\renewcommand\arraystretch{0.9}
	\begin{tabular}{p{2.8cm}<{\centering}|p{0.9cm}<{\centering}|p{0.9cm}<{\centering}|p{0.9cm}<{\centering}|p{0.9cm}<{\centering}|p{0.9cm}<{\centering}|p{0.9cm}<{\centering}|p{0.9cm}<{\centering}|p{0.9cm}<{\centering}|p{0.9cm}<{\centering}|p{0.9cm}<{\centering}|p{0.9cm}<{\centering}}
		\toprule
		Dataset & \multicolumn{3}{c|}{ACM} & \multicolumn{2}{c|}{DBLP} & \multicolumn{4}{c|}{Douban} & \multicolumn{2}{c}{Yelp} \\
		\midrule
		Node Type & Paper & Author & Venue & Paper & Author & User & Movie & Director & Actor & User & Business \\
		\midrule

		DeepWalk & 
		0.1983 & 0.2403 & 0.3705 & 0.6504 & 0.9175 & 0.1789 & 0.3136 & 0.2408 & 0.2661 & 0.0584 & 0.1187 \\
		LINE & 
		0.1891 & 0.2358 & 0.3649 & 0.6863 & 0.9238 & 0.2064 & 0.3120 & 0.2221 & 0.2382 & 0.0389 & 0.1164 \\
		PTE &
		0.1941 & 0.2474 & 0.3780 & 0.6376 & 0.9218 & 0.1687 & 0.2962 & 0.2279 & 0.2405 & 0.0314 & 0.1065 \\
		\midrule
		Metapath2vec-$\mathcal P_{1}$ & 
		0.1682 & 0.2408 & 0.4239 & 0.8073 & 0.9344 & 0.3703 & 0.2871 & -- & 0.2732 & 0.0683 & 0.1147 \\
		Metapath2vec-$\mathcal P_{2}$ & 
		0.2019 & 0.3174 & -- & 0.5110 & 0.8605 & 0.3288 & 0.2537 & 0.2305 & -- & 0.0782 & 0.0609 \\ 
		\hline	
		Metagraph2vec-$\mathcal S_{1}$ &
		0.1757 & 0.2973 & 0.4398 & 0.7988 & 0.9309 & 0.3958 & 0.3117 & 0.2571 & 0.2767 & 0.1362 & 0.1337 \\ 
		Metagraph2vec-$\mathcal S_{2}$ &
		0.2113 & 0.3080 & 0.4254 & 0.7832 & 0.9462 & 0.3866 & 0.2606 & 0.2176 & 0.2351 & 0.1217 & 0.1263 \\ 
		\midrule
		SpaceyMetapath-$\mathcal P_{1}$ &
		0.1797 & 0.2817 & 0.4518 & 0.8329 & 0.9399 & 0.3827 & 0.2921 & -- & 0.3101 & 0.1297 & 0.1310 \\ 
		SpaceyMetapath-$\mathcal P_{2}$ & 
		0.2123 & 0.3571 & -- & 0.5506 & 0.8774 & 0.3613 & 0.2767 & 0.2615 & -- & 0.0968 & 0.1089 \\ 
		\hline
		SpaceyMetagraph-$\mathcal S_{1}$ & 
		0.1862 & 0.3290 & 0.4616 & {\bf 0.8361} & 0.9363 & {\bf  0.4131} & {\bf 0.3154} & {\bf 0.2859} & {\bf 0.3247} & {\bf 0.1476} & {\bf 0.1431} \\ 
		SpaceyMetagraph-$\mathcal S_{2}$ & 
		0.2188 & 0.3511 & 0.4629 & 0.8243 & {\bf 0.9485} & 0.4015 & 0.2890 & 0.2678 & 0.3030 & 0.1270 & 0.1315 \\ 
		\hline
		SpaceyMetaschema &
		{\bf 0.2304} & {\bf 0.3587} & {\bf 0.4641} & 0.8196 & 0.9478 & 0.3827 & 0.3012 & 0.2699 & 0.3120 & 0.0768 & 0.1268 \\ 
		\bottomrule
	\end{tabular}
	\vspace{-0.1in}
\end{table*}

\subsection{Multi-label Node Classification}

For the node classification task, 
we first learn the node embedding vectors from the full nodes on each dataset, and then use the embeddings of the labeled nodes as input features for a one-vs-rest logistic regression classifier. 
We repeat each classification experiment ten times by randomly splitting 50\% of the labeled nodes for training and the others for testing, and report the average performance in terms of both Micro-F1 score and Macro-F1 score.
We also calculate the variances of the scores to reflect the significance of the experimental results.

We can observe that our proposed algorithms consistently outperform all the start-of-the-art baselines in both metrics on all datasets.
For example, 
for the author node classification on the ACM dataset,
our proposed models outperform all baseline models by around 0.01--0.10 (relatively 1\%--18\%) in terms of Micro-F1 scores and by around 0.04--0.12 (relatively 13\%--52\%) in terms of Macro-F1 scores.
Moreover, given the same meta-path or meta-graph, the proposed {\it SpaceyMetapath}/{\it SpaceyMetagraph} imporves the author node classification performance by around 0.01--0.03 in Micro-F1 score and 0.03--0.04 in Macro-F1 score over Metapath2vec/Metagraph2vec.

In most cases, we can find that,
by capturing the semantically meaningful information of an appropriate meta-path or meta-graph,
the meta-path/meta-graph driven algorithms (including our proposed methods and Metapath2vec and Metagraph2vec) obviously outperform DeepWalk, LINE and PTE.
Among them, meta-graph driven algorithms can achieve better performance than meta-path driven algorithms by capturing richer semantics of integrating multiple meta-paths. 
In addition, 
we can find that
the proposed {\it SpaceyMetaschema} achieves competitive performance compared with the best of other methods,
which may be more practical in complex networks with a large variety of types for the dispense with handcrafted domain-specific meta-paths/meta-graphs.

\begin{table}[!htb]
	\small
	\caption{\small Binary operators for learning edge features. The definitions correspond to the $i$-th component of vector ${\mathbf u}$ and ${\mathbf v}$.}\label{Tab:binary_operators}
	\vspace{-0.15in}
	\centering
	\renewcommand\arraystretch{0.7}
	\begin{tabular}{c|c|c|c|c}
		\toprule
		Operator & Average & Hadamard & Weighted-L1 & Weighted-L2 \\
		\midrule
		Definition & $({\mathbf u}_i+{\mathbf v}_i) / 2$ & ${\mathbf u}_i * {\mathbf v}_i$ & $|{\mathbf u}_i - {\mathbf v}_i|$ & ${|{\mathbf u}_i - {\mathbf v}_i|}^2$ \\
		\bottomrule
	\end{tabular}
	\vspace{-0.15in}
\end{table}

\begin{table*}[t]
	 	\small
	\caption{\small The mean AUC scores for link prediction. ``--'' indicates that the used meta-path/meta-graph cannot generate embeddings for nodes of target type. The edge types for each dataset are shown in Figure~\ref{Fig:datasets}.}\label{Tab:lp_results}
	\vspace{-0.15in}
	\centering 
	\renewcommand\arraystretch{0.9}
	\begin{tabular}{p{2.8cm}<{\centering}|p{0.8cm}<{\centering}|p{0.8cm}<{\centering}|p{0.8cm}<{\centering}|p{0.8cm}<{\centering}|p{0.8cm}<{\centering}|p{0.8cm}<{\centering}|p{0.8cm}<{\centering}|p{0.8cm}<{\centering}|p{0.8cm}<{\centering}|p{0.8cm}<{\centering}|p{0.8cm}<{\centering}|p{0.8cm}<{\centering}}
		\toprule
		Dataset & \multicolumn{3}{c|}{ACM} & \multicolumn{3}{c|}{DBLP} & \multicolumn{3}{c|}{Douban} & \multicolumn{3}{c}{Yelp} \\
		\midrule
		Edge Type & P--P & P--V & P--T & P--A & P--C & P--T & U--U & U--M & M--D & U--U & U--B & B--C \\
		\midrule
		DeepWalk &
		0.8711 & 0.8775 & 0.6640 & 0.8743 & 0.9219 & 0.8777 & 0.6255 & 0.7906 & 0.8397 & 0.8768 & 0.8815 & 0.9212 \\ 
		LINE &
		0.8555 & 0.8550 & 0.6454 & 0.8180 & 0.8550 & 0.7479 & 0.6374 & 0.7850 & 0.8069 & 0.8770 & 0.8545 & 0.9063 \\ 
		PTE &
		0.6734 & 0.7189 & 0.5776 & 0.6410 & 0.7189 & 0.5901 & 0.6691 & 0.6461 & 0.5881 & 0.8319 & 0.7969 & 0.8244 \\\midrule
		Metapath2vec-$\mathcal P_{1}$ &
		0.7137 & 0.9486 & -- & 0.8801 & 0.9532 & -- & 0.6061 & 0.7994 & -- & 0.6464 & 0.8382 & 0.9179 \\
		Metapath2vec-$\mathcal P_{2}$ &
		0.7392 & -- & 0.7307 & 0.8848 & -- & 0.8900 & 0.6006 & 0.8099 & 0.8793 & 0.7333 & 0.7841 & -- \\\hline
		Metagraph2vec-$\mathcal S_{1}$ &
		0.7257 & 0.9393 & -- & 0.8887 & 0.9438 & -- & 0.6097 & 0.8081 & 0.8907 & 0.6493 & 0.8664 & 0.9081 \\ 
		Metagraph2vec-$\mathcal S_{2}$ &
		0.7313 & 0.9467 & 0.6865 & 0.8925 & 0.9555 & 0.8765 & 0.6728 & 0.7711 & 0.8624 & 0.6919 & 0.8718 & -- \\\midrule
		SpaceyMetapath-$\mathcal P_{1}$ &
		0.7236 & 0.9521 & -- & 0.8901 & 0.9596 & -- & 0.6206 & 0.8318 & -- & 0.6668 & 0.8565 & 0.9371 \\
		SpaceyMetapath-$\mathcal P_{2}$ &
		0.7432 & -- & {\bf 0.7365} & 0.8870 & -- & {\bf 0.8926} & 0.6152 & 0.8310 & 0.8976 & 0.7507 & 0.8516 & -- \\\hline
		SpaceyMetagraph-$\mathcal S_{1}$ &
		0.7292 & 0.9427 & -- & 0.8911 & 0.9511 & -- & 0.6240 & 0.8339 & {\bf 0.9022} & 0.6537 & 0.8848 & 0.9242 \\
		SpaceyMetagraph-$\mathcal S_{2}$ &
		0.7333 & {\bf 0.9523} & 0.6889 & {\bf 0.8956} & 0.9618 & 0.8830 & 0.6903 & 0.8011 & 0.9009 & 0.6976 & 0.8868 & -- \\\hline
		SpaceyMetaschema &
		{\bf 0.9238} & 0.9385 & 0.6731 & 0.8867 & {\bf 0.9655} & 0.8824 & {\bf 0.9338} & {\bf 0.8734} & 0.9015 & {\bf 0.9254} & {\bf 0.8970} & {\bf 0.9497} \\
		\bottomrule
	\end{tabular}
	\vspace{-0.1in}
\end{table*}

\subsection{Link Prediction}
For each dataset, 
we follow \cite{GroverL16} to randomly hide 20\% of edges of each edge type as missing edges.
Then, we learn the embedding using the rest of the 80\% edges and predict these missing edges.
For the computational effciency, we randomly sample 2,048 edges (from the 20\% hided edges) as positive examples and equally split them into two partitions $E_{train}$ and $E_{test}$.
We also randomly sample 2,048 unobserved edges as negative examples and equally split them into two partitions $E_{train}^-$ and $E_{test}^-$.
Then, we consider the link prediction evaluation as a binary classification problem with  $(E_{train}, E_{train}^-)$ for training and  $(E_{test}, E_{test}^-)$ for testing.
We study several binary operators~\cite{GroverL16} (as shown in Table~\ref{Tab:binary_operators}) to construct features for an edge based on its two node vectors, and then the operated feature vector for an edge is as the input to a logistic regression classifier.
We repeat the above link prediction experiment ten times and report the average performance in terms of AUC (Area Under Curve) score.

Overall, the results of link prediction are consistent with the results of node classification, and we can reach a similar conclusion as analyzed in node classification experiment.
We can see that the proposed methods clearly achieve better performance than the comparative methods on all datasets.
For example, 
on the ACM dataset,
we improve the link prediction performance of different edge types by around 0.05--0.25 (relatively 6\%--37\%) over DeepWalk, LINE and PTE, and 0.01--0.21 (relatively 1\%--29\%) over Metapath2vec and Metagraph2vec, in terms of the mean AUC score.
For the detailed AUC scores of all binary operators (which are reported in the supplementary material), the improvements are consistent.
For example, on the DBLP dataset,
we can observe that the proposed methods achieve 0.01--0.09 (relatively 1\%--10\%) gains over all baseline methods in all the link prediction experiments of different edge types, in terms of the best possible choice of the binary operators for each algorithm.
Moreover,
we can also evidently observe the effectiveness of the proposed spacey random walk compared with the normal meta-path/meta-graph based random walk.
In most cases, given the same meta-path or meta-graph, the proposed {\it SpaceyMetapath}/{\it SpaceyMetagraph} can achieve around 0.01--0.07 improvements over Metapath2vec/Metagraph2vec in the link prediction experiments of different edge types on different datasets.

It is worth mentioning that, in some cases,
the {\it SpaceyMetaschema} significantly outperforms {\it SpaceyMetapath}/{\it SpaceyMetagraph}, e.g., for the P–P link prediction on the ACM dataset and the U-U link prediction on the Yelp dataset, the improvements are almost 0.2.
And even, in these cases, the homogeneous methods (e.g., DeepWalk and LINE ) can also perform better than the meta-path/meta-graph based methods.
The possible reason we believe is the specialities of the relations and the biases of the used meta-paths/meta-graphs.
A given meta-path/meta-graph usually expresses some important semantics, but may not be very sufficient to reflect other special link structures.
As a result, the methods which homogeneously cover all links (i.e., DeepWalk and LINE) and which dynamically treat all links with a superior spacey stragety (i.e.,  {\it SpaceyMetaschema}) abnormally achieve better performance in these cases.


\begin{figure*}[t]
	\small
	\setlength{\abovecaptionskip}{-0.1cm}
	\setlength{\belowcaptionskip}{-0.4cm}
	\centering
	\subfigure[\small walk-times $w$.]{
		\label{Fig:walk_times_P1}
		\includegraphics[width=0.3\textwidth]{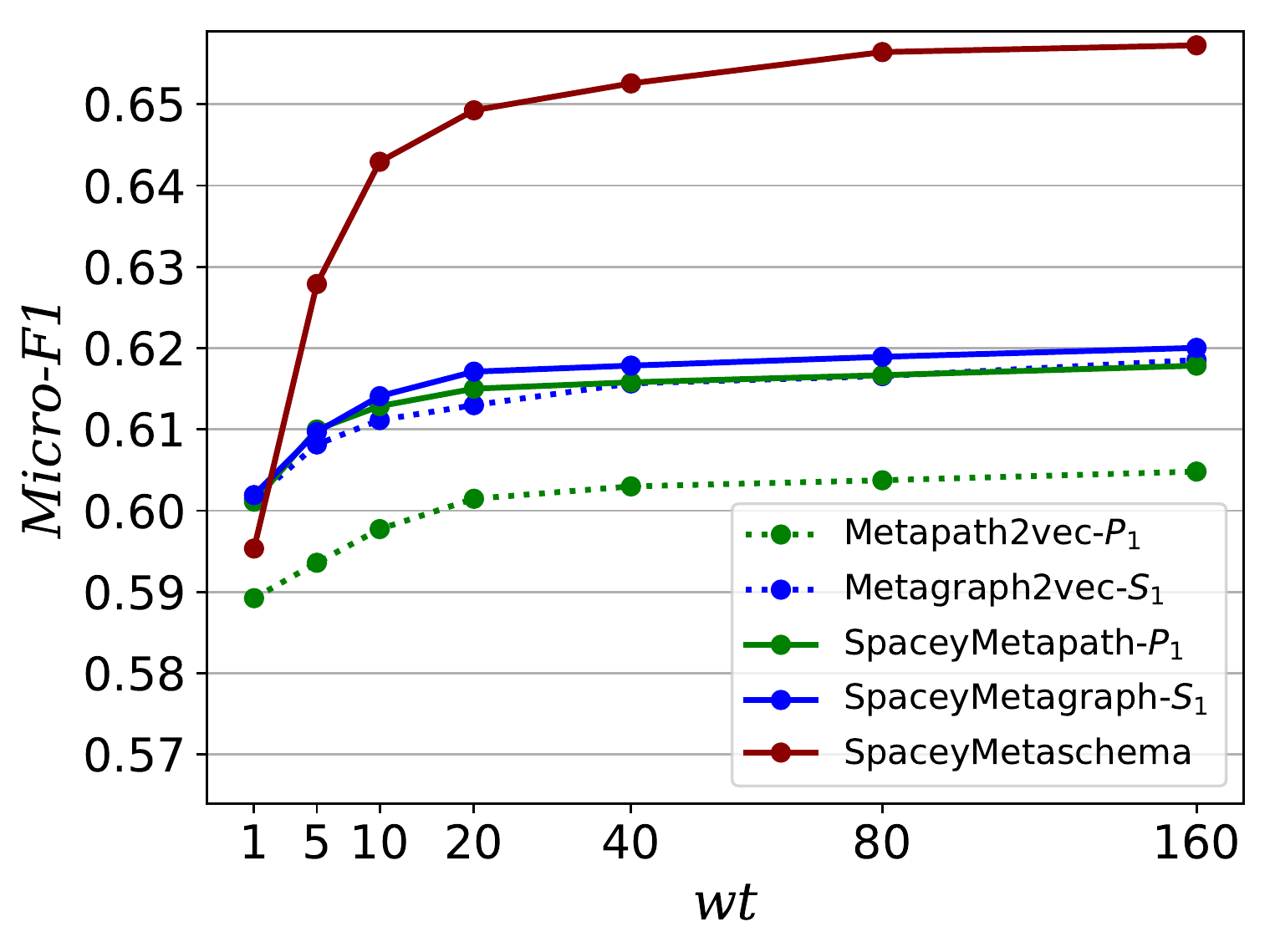}
	}
	\subfigure[\small walk-length $l$.]{
		\label{Fig:walk_length_P1}
		\includegraphics[width=0.3\textwidth]{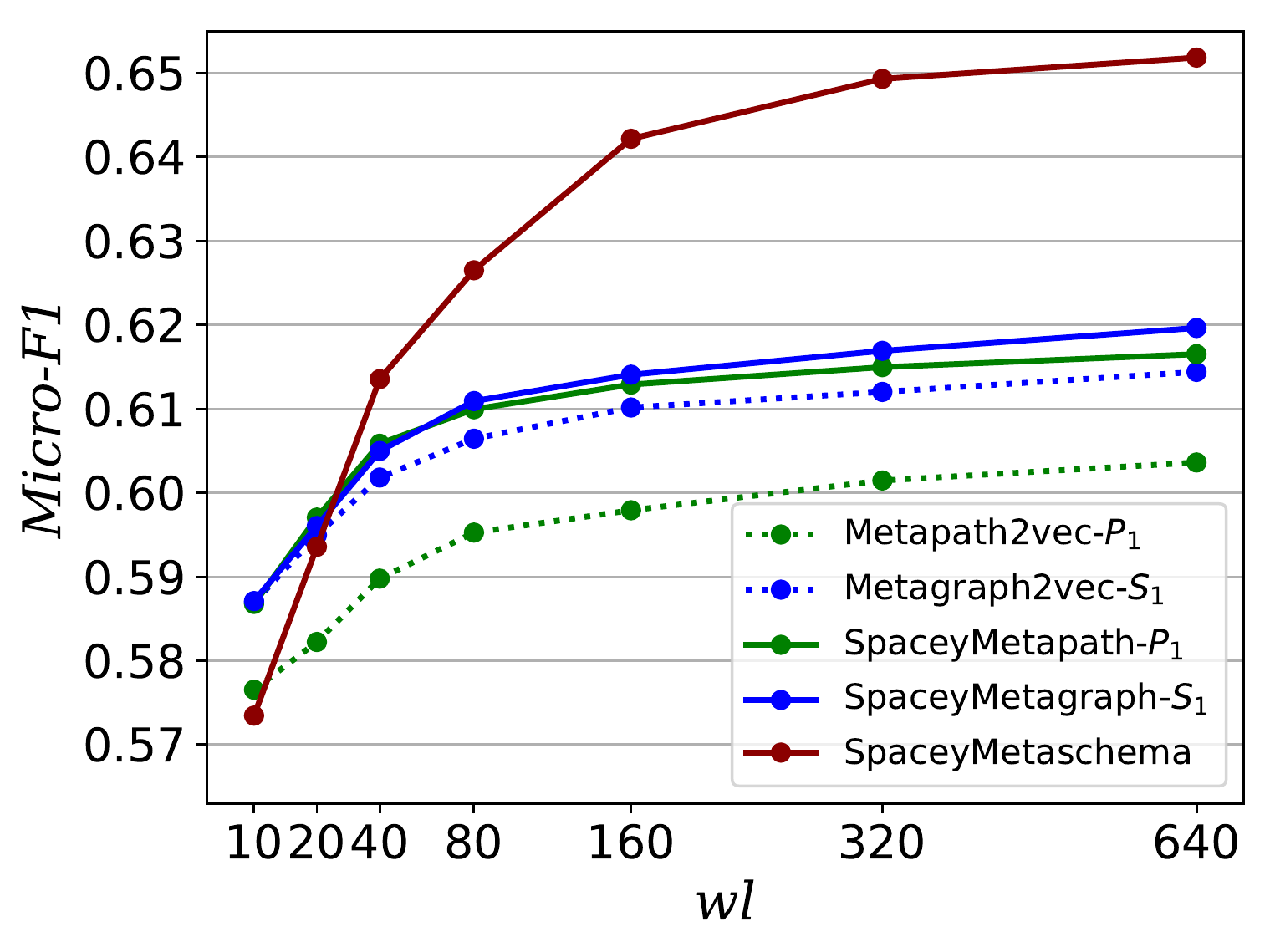}
	}
	\subfigure[\small personalized-probability $\alpha$.]
	{\label{Fig:personalized_probability_P1}
		\includegraphics[width=0.3\textwidth]{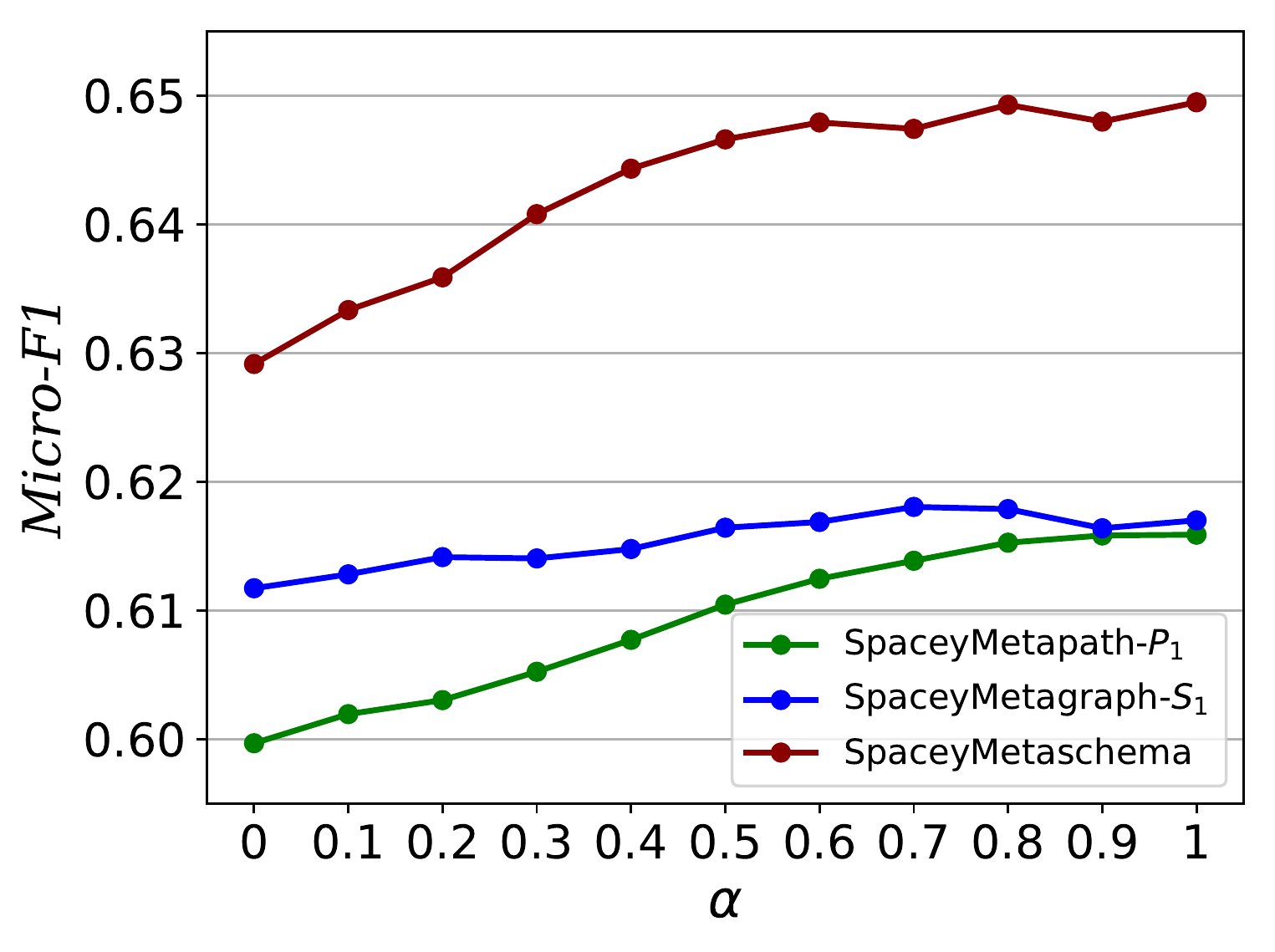}
	}
	\caption{\small Parameter sensitivity for author node classification on the ACM dataset.}
	\label{Fig:Parameter_sensitivity}
	
\end{figure*}

\vspace{-0.1in}
\subsection{Parameter Sensitivity}
\label{sec:Parameter_Sensitivity}
In this section, we illustrate the parameters sensitivity 
by the Micro-F1 scores of author node classification experiments on the ACM dataset. 
For each experiment, we vary one parameter and fix the others as the default values (as shown in Sec.~\ref{sec:exp_settings}).

\begin{figure}[t]
	\small
	\setlength{\abovecaptionskip}{-0.1cm}
	\setlength{\belowcaptionskip}{-0.4cm}
	\centering
	\includegraphics[width=0.37\textwidth]{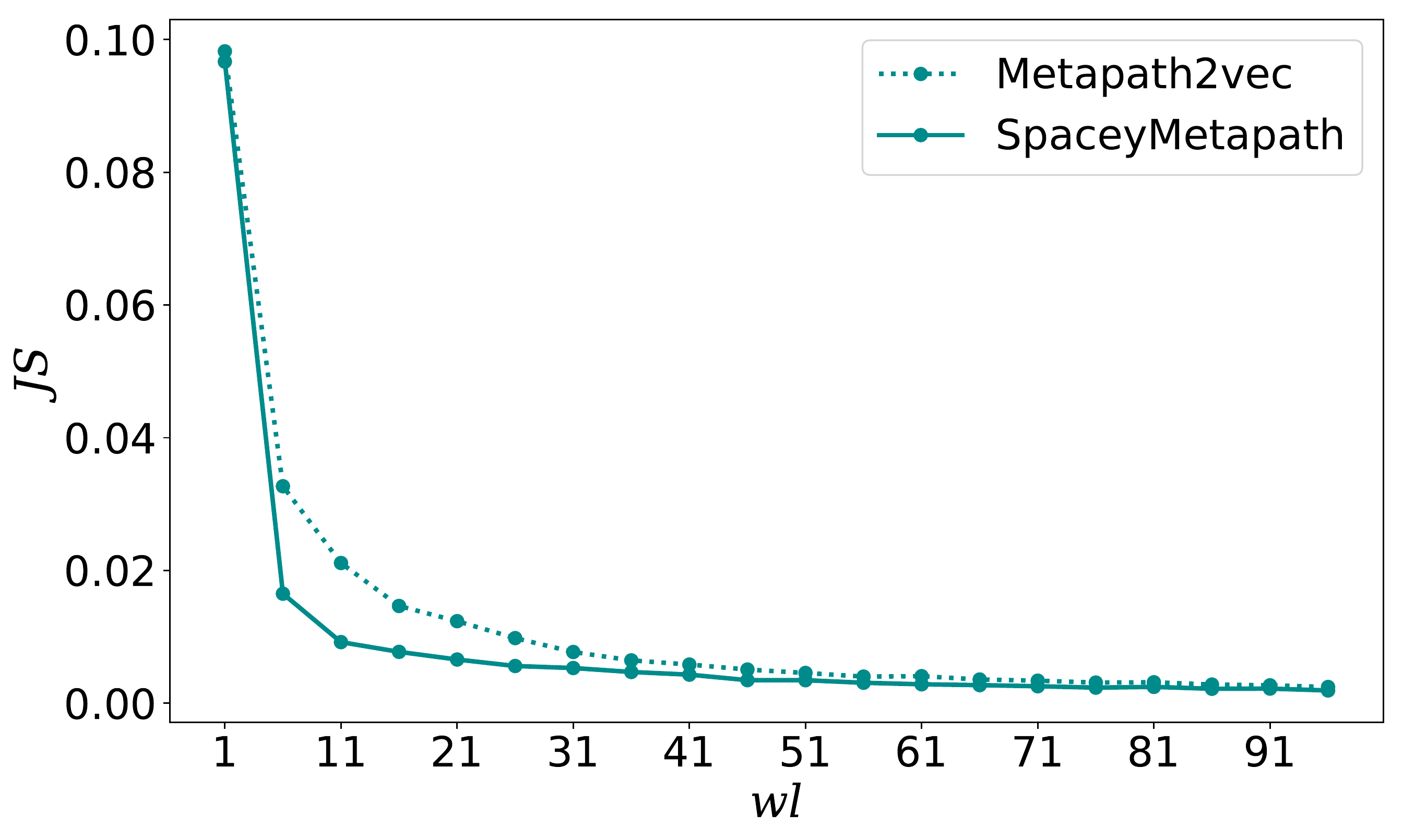}
	\caption{\small Variation of node distribution during meta-path ``APVPA'' guided random walks on the ACM dataset.}
	\label{Fig:stationary_analysis_P1}
\end{figure}

From Figure~\ref{Fig:walk_times_P1} and Figure~\ref{Fig:walk_length_P1}, we can see that walk-times $w$ and walk-length $l$ are positive to 
performance for all random walk based algorithms, and the positivity gradually weakens with the parameter increasing.
From Figure~\ref{Fig:personalized_probability_P1}, we can see that the performance increases with the  personalized probability $\alpha$ increasing, and the tend will be weakened when $\alpha$ reaches around 0.7.
Specially, from Figure~\ref{Fig:Parameter_sensitivity},  
we can clearly observe that the proposed {\it SpaceyMetapath}/{\it SpaceyMetagraph} outperforms Metapath2vec/Metagraph2vec given a relatively small $w$ or $l$.
For example,
given meta-path $\mathcal P_1$ as the input, we can find that the performance of the proposed {\it SpaceyMetapath} with $l\!=\!40$ is equivalent to the performance of Metapath2vec with $l\!=\!640$, giving us around 16x speedup.
Further,
we randomly trace 1000 walk sequences of meta-path $\mathcal P_1$ guided spacey random walk and normal Marovian random walk respectively,
and gradually compute the JS (Jensen--Shannon) divergence~\cite{fuglede2004jensen} of the node distributions between steps.
The average variation is shown in Figure~\ref{Fig:stationary_analysis_P1}.
Compared with Metapath2vec, 
we can evidently observe that the distribution divergence for {\it SpaceyMetapath} quickly descends to a small level with a small walk-length.
Overall, above comparision analysis indicates that the proposed spacey random walk frameworks are able to reach high performance under a cost-effective parameter choice (the smaller, the more efficient).


\subsection{Complexity and Scalability Analysis}
\begin{figure}[t]
	\small
	\setlength{\abovecaptionskip}{-0.1cm}
	\setlength{\belowcaptionskip}{-0.4cm}
	\centering
	\includegraphics[width=0.45\textwidth]{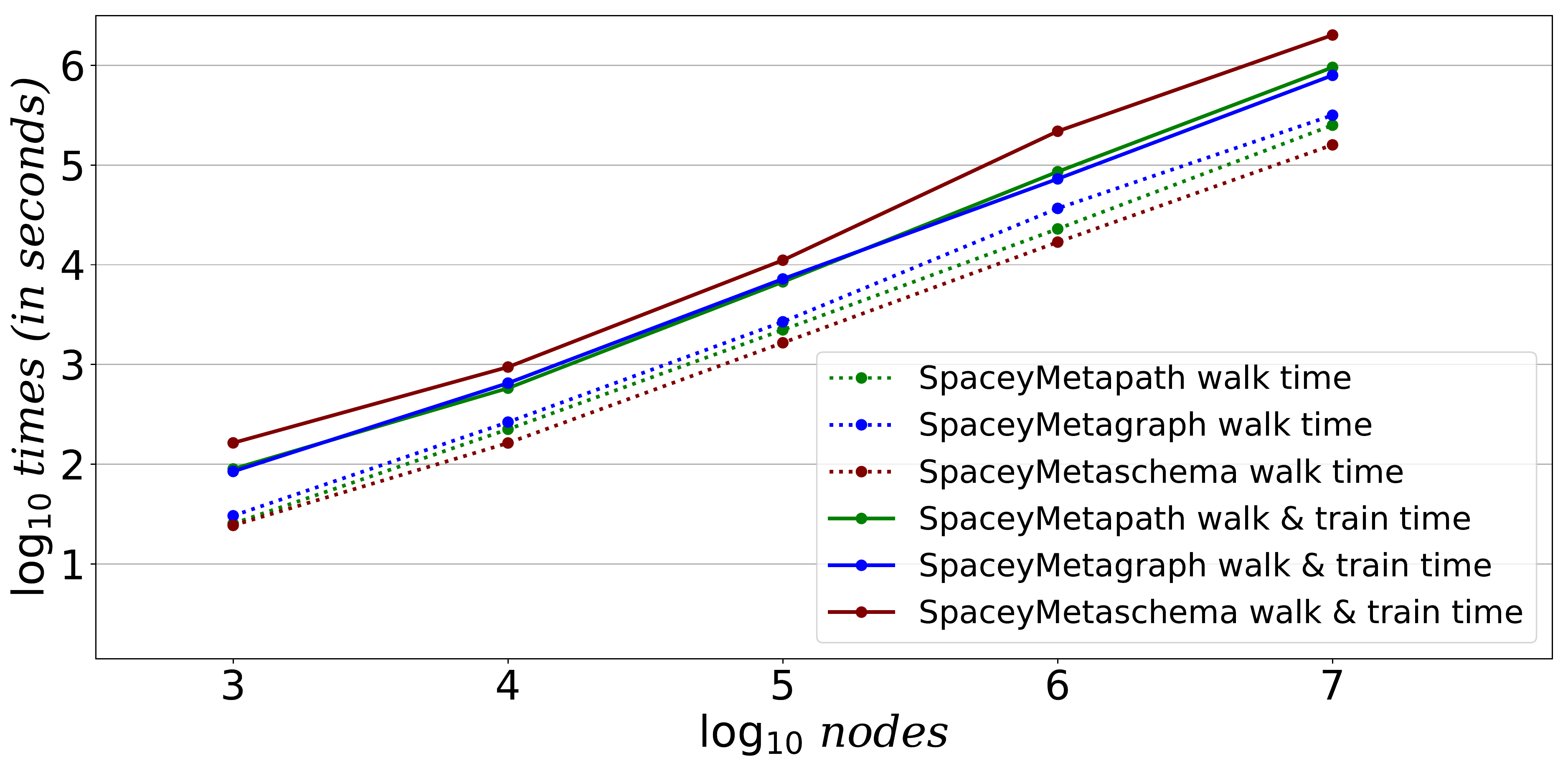}
	\caption{\small The walk and train time of the proposed models over different network sizes.}
	\label{Fig:scalability_time}
\end{figure}

In this section, we demonstrate the scalability of the proposed frameworks by measuring both the walk time to generate heterogeneous neighborhood and the training time to learn node embeddings.
We first follow the meta-schema of DBLP network to simulate a series of random graph datasets with the average degree of 10.
The specific network sizes are [1k; 10k; 100k; 1000k; 10000k].
Then, we independently run experiments on these synthetic HINs in a computing server with $56$ core Intel(R) Xeon(R) CPU E5-2680 v4 @ 2.40GHz. 
The time consumption is shown in Figure~\ref{Fig:scalability_time}, which shows that our methods all have a linear time complexity with respect to the number of nodes and can be easily applied to very large-scale networks.
In fact, 
for the random walk phase in an HIN ${\mathcal G} = ({\mathcal V}, {\mathcal E}; \mathcal A, \mathcal R)$ with  walk-times $t$ and walk-length $l$,
the spacey walk for each step takes $O(|\mathcal A|)$ time, and the total walk complexity is $O(wl|\mathcal A||\mathcal V|)$, which is linear to the number of nodes.
For the training phase, we use a Skipgram model to train the embeddings of different types of nodes, which also has a linear complexity  and can be parallelized by using the same mechanism as word2vec~\cite{GroverL16}.
Overall, the proposed algorithms are quite efficient and scalable for large-scale heterogeneous networks.

\section{Related Work}\label{sec:related-work}
In the past decade,
to marry the advantages of HIN and network embedding, 
embedding learning in an HIN has received increasing attention and many heterogeneous embedding algorithms have been proposed~\cite{Bordes2013,Tang2015PTE,chang2015heterogeneous,YuxiaoDong2017,zhang2018carl,kralj2018hinmine,qiu2018network,sun2018joint,shi2018easing}.

In general,
there have been multiple ways to represent multi-typed nodes in an HIN.
A straightforward method is to directly perform matrix/tensor factorization~\cite{MaruhashiGF12,PapalexakisFS17,qiu2018network} to extract vector representations of nodes. 
For example, 
{\it Collective matrix factorization} of multi-type relational data~\cite{Singh2008,NickelTK12,BouchardYG13} treats each binary relation as a matrix, and performs joint matrix factorization over different relations.
However, such a way is usually costly for computation and memory.
More recently, researchers apply stochastic optimization to optimize simple or deep models to predict the binary relations between two types of entities with large-scale knowledge graphs~\cite{Nickel0TG16}.
Representative approach include {TransE}~\cite{Bordes2013} where one entity in a type is translated by a relation vector to another entity in another type given the co-occurrence of a binary relations between them and PTE~\cite{Tang2015PTE} which decomposes an HIN to a set of bipartite networks by edge types, and learns node vectors by capturing 1-hop neighborhood of the resulting bipartite networks.
All of these methods, although sometimes called higher-order factorization or embedding, are still dealing with the co-occurrence of binary relations among entities~\cite{Nickel0TG16}.
More recently, inspired by the DeepWalk model for homogeneous networks~\cite{PerozziAS14}, meta-paths or meta-graphs guided random walk based entity embedding models have also been well developed~\cite{shang2016meta,YuxiaoDong2017,FuLL17,ChuanShi2017,zhang2018metagraph2vec} using the Skipgram techniques introduced by word2vec~\cite{Mikolov2013} for very large graphs.
However, as discussed in Section~\ref{sec:introduction}, 
these algorithms have few attention on the higher-order Markov chain nature of meta-path guided random walk.
Following random walk based embedding algorithms, we focus on the higher-order Markov chains and design a new meta-path/meta-graph guided random walk strategy, and further develop several general scalable HIN embedding models.

\section{Conclusions}\label{sec:conclu}

In this paper, we propose a heterogeneous personalized spacey random walk strategy to efficiently generate heterogeneous neighborhood, which is based on a solid mathematical foundation. We further develop a scalable HIN embedding algorithms {\it SpaceyMetapath} to efficiently and effectively produce node embeddings in an HIN guided by an arbitrary meta-path.
We also develop two scalable HIN embedding algorithms by extending the {\it SpaceyMetapath} from a meta-path to a meta-graph or a meta-schema.
Experimental results show that our methods are able to achieve better performance with smaller walk-times and walk-length.

\section{Acknowledgments}
This work is supported by the China NSFC program (No.61872022, 61421003) , SKLSDE-2018ZX16, and the Early Career Scheme (ECS, No.26206717) from Research 
Grants Council in Hong Kong. For any correspondence, please refer to Jianxin Li.

\bibliographystyle{ACM-Reference-Format}
\bibliography{sigproc}

\end{document}